\documentclass[10pt,twocolumn]{article}
\usepackage[letterpaper,margin=0.80in]{geometry}
\usepackage[round]{natbib}

\usepackage{amsmath,amssymb}
\usepackage{amsthm}
\newtheorem{theorem}{Theorem}
\newtheorem{proposition}{Proposition}

\usepackage{graphicx}
\usepackage{booktabs}
\usepackage{placeins}
\usepackage{xcolor}
\usepackage{float}
\usepackage{multirow}
\usepackage{tabularx}
\usepackage{tikz}
\usetikzlibrary{arrows.meta,positioning,fit}

\definecolor{bestblue}{RGB}{0,114,178}
\definecolor{secondorange}{RGB}{230,159,0}
\newcommand{\best}[1]{\textcolor{bestblue}{\textbf{#1}}}
\newcommand{\second}[1]{\textcolor{secondorange}{#1}}

\definecolor{theoryamber}{RGB}{178,118,35}
\definecolor{empblue}{RGB}{45,95,145}
\definecolor{abstainred}{RGB}{165,70,70}
\newcommand{\theory}[1]{\textcolor{theoryamber}{#1}}
\newcommand{\empirical}[1]{\textcolor{empblue}{#1}}
\newcommand{\abstain}[1]{\textcolor{abstainred}{\textbf{#1}}}
\graphicspath{{figures/}}

\begin{document}
\twocolumn[{
\begin{center}
{\LARGE\bfseries When Should Semantic Priors Enter\\[2pt]
Partially-Identified Causal Discovery?\par}
\vspace{16pt}
\noindent
\begin{minipage}[t]{0.31\textwidth}
\centering
\textbf{Muhammad Ahmed Mohsin}\\[3pt]
{\small Stanford University}
\end{minipage}\hfill
\begin{minipage}[t]{0.31\textwidth}
\centering
\textbf{Saad Bin Ather}\\[3pt]
{\small FAST-NUCES}\\
{\small Lahore, Pakistan}
\end{minipage}\hfill
\begin{minipage}[t]{0.31\textwidth}
\centering
\textbf{Valentin Six}\\[3pt]
{\small University of Virginia}\\
{\small Charlottesville, USA}
\end{minipage}

\vspace{16pt}
\noindent
\begin{minipage}[t]{0.31\textwidth}
\centering
\textbf{Lancelot Da Costa}\\[3pt]
{\small VERSES Technologies Inc.}
\end{minipage}\hfill
\begin{minipage}[t]{0.31\textwidth}
\centering
\textbf{Tommaso Salvatori}\\[3pt]
{\small Relational Intelligence}
\end{minipage}\hfill
\begin{minipage}[t]{0.31\textwidth}
\centering
\textbf{Bernhard Sch\"olkopf}\\[3pt]
{\small ELLIS Institute}
\end{minipage}
\end{center}
\vspace{16pt}
}]

\begin{abstract}
Observational causal discovery often leaves causal directions unresolved
or unstable. Large language models (LLMs) can provide semantic information
to resolve these directions, but incorrect proposals can exclude the true
causal graph and invalidate downstream inference. We study three separate
decisions: where to allocate semantic queries, which proposals to admit,
and which causal graphs to retain. Our framework distinguishes population
non-identifiability from finite-sample instability, evaluates semantic
orientation against statistical and random controls, and examines
calibrated admission and error-tolerant DAG retention. Experiments across
benchmark networks show that semantic information improves recovery over
coin-flip orientations, but targeting unstable edges has no established
advantage over matched-random allocation. The tested admission certificate
cannot support nontrivial admission with the available calibration tasks.
Moreover, downstream retention depends on reference adequacy: learned
references contain the true DAG in only three of 80 planned tasks.
These findings separate the value of semantic information from the
reliability of its allocation, admission, and retention, and identify
calibration size and reference adequacy as key limitations.
\end{abstract}

\section{Introduction}
\label{sec:introduction}

Observational causal discovery aims to recover causal structure from
nonexperimental data. Causal graphs support intervention analysis,
effect estimation, and downstream decisions. However, observational data
often cannot determine every causal direction. Different DAGs can explain
the same distribution, while finite samples can make directions unstable.
A fitted graph can therefore contain unresolved or unreliable orientations.

\begin{figure}[H]
\centering
\includegraphics[width=\columnwidth]
{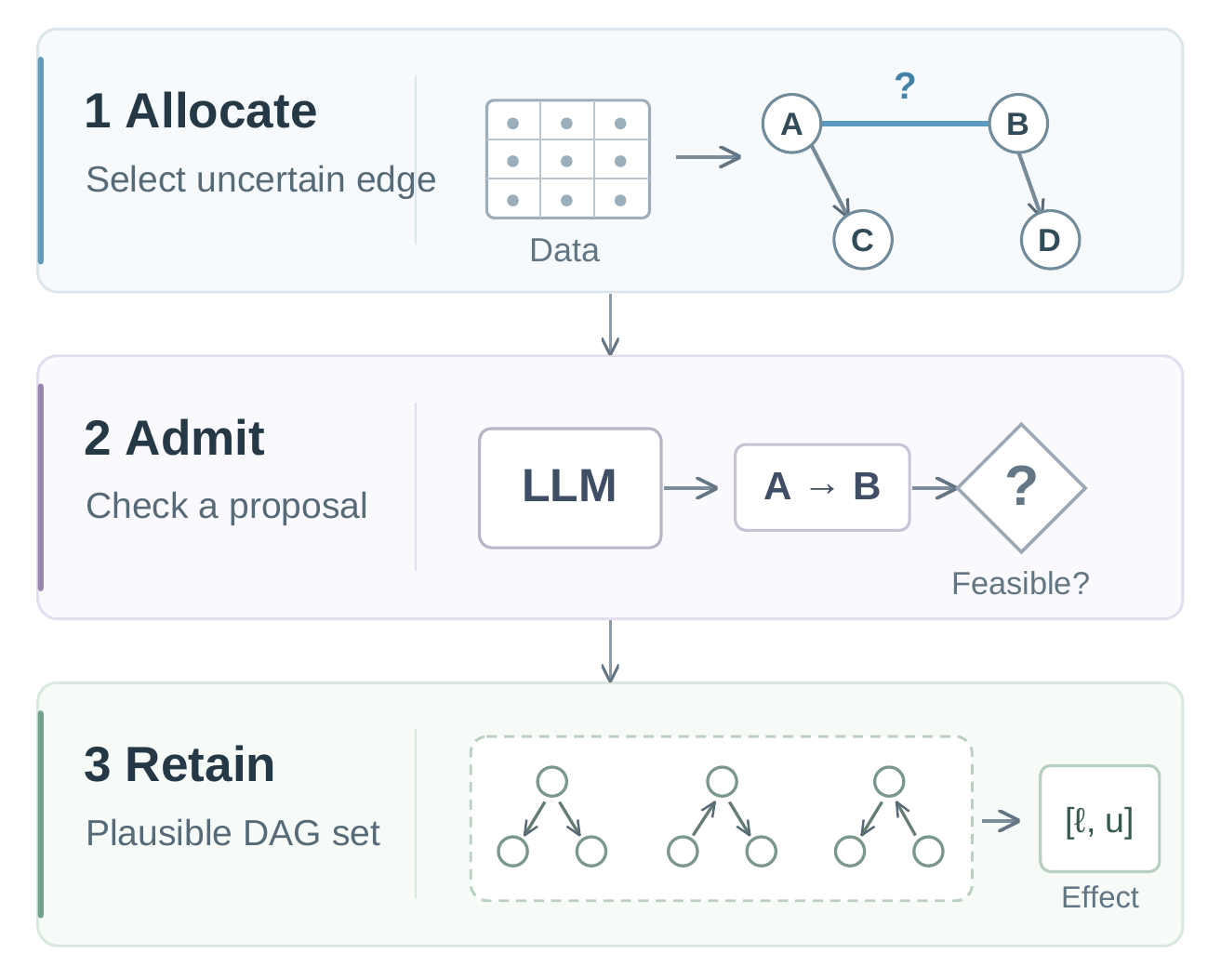}
\caption{\textbf{Semantic assistance in causal discovery.}
Allocation selects queries, admission checks proposals, and
retention preserves plausible DAGs.}
\label{fig:motivation}
\end{figure}

Semantic knowledge can provide information absent from observational data. 
Variable names and descriptions can encode mechanisms, temporal order,
and domain relationships. Large language models can use these cues to 
propose causal directions
\citep{kiciman2023causal,ban2023query,vashishtha2023causal,long2023causal}.
However, semantic advice is fallible. A wrong orientation can exclude the
true causal graph and change downstream conclusions. We therefore ask
where semantic information should be queried, when it should be admitted,
and how much graph uncertainty should remain.

We separate these decisions into \emph{allocation}, \emph{admission}, and
\emph{retention}. Allocation selects query targets. Admission determines
which proposals may alter the graph. Retention preserves plausible causal
models for downstream inference. Figure~\ref{fig:motivation} illustrates
these stages. Improved graph recovery does not imply calibrated admission
risk, and low admission risk does not guarantee truth retention.

We distinguish two sources of directional uncertainty for query allocation.
Some directions are unidentified under the assumed model class. Others are
population-identified but unstable at finite sample sizes. We call these
regions F1 and F2, respectively. Their fitted counterparts use unresolved
edges and bootstrap-unstable committed arrows as operational candidates.
Across the four primary estimator--sample-size settings, $36.5$--$49.0\%$
of fitted skeleton edges fall into these candidate regions
(see Figure~\ref{fig:edge-composition} in Appendix~\ref{app:recovery}).
These fitted categories indicate where semantic information may help,
but do not establish causal error or population non-identifiability.

Our experiments test the three decisions separately. The tested F2 locator
does not consistently outperform matched-random targeting, and all four
primary allocation intervals include zero. Once an uncertain direction is
selected, semantic resolution can help. On three additional benchmark
families, semantic resolution improves Directed-F1 over bootstrap-only
resolution by $0.42$ percentage points (pp), with exploratory interval
$[0.24,0.59]$ pp. A bootstrap--LLM agreement gate and a propagation-based
allocation rule also fail to dominate across estimator regimes.
These results separate semantic content from query-location quality.

Admission and retention impose different constraints. A finite-grid
Hoeffding certificate cannot support nontrivial admission from our
available calibration pools. This limitation concerns the tested
certificate and available independent graph tasks, not a universal
sample-complexity barrier. For retention, we calibrate an allowed number
of semantic disagreements and keep all stored DAGs within that allowance.
This guarantee depends on reference adequacy. In the learned-reference
study, the stored reference contains the true DAG in only three of 80
planned tasks.

We make three contributions. First, we separate semantic information,
query allocation, and directional resolution through coin-flip,
matched-random, bootstrap-only, agreement-gated, and propagation-based
controls. We also provide a native PC-family comparison to Guess2Graph.
Second, we separate admission risk from downstream retention and connect
both to explicit graph-task and reference-adequacy assumptions. Third, we
evaluate these stages through semantic controls, certification diagnostics,
fresh Oracle draws, and learned references. The concentration and
rank-calibration arguments use standard tools. Our contribution is their
operational separation and the evidence that different stages succeed or
fail for different reasons.

\section{Related Work}
\label{sec:related}

\paragraph{Discovery and background knowledge.}
Constraint-based discovery uses conditional independence and orientation
rules \citep{spirtes2000causation,meek2013causal}. Score-based discovery
searches over structures or equivalence classes
\citep{chickering2002optimal}. Identifiability depends on the model class.
Equal error variances and suitable additive-noise restrictions can identify
directions beyond an unrestricted linear-Gaussian equivalence class
\citep{peters2014identifiability,peters2014causal}.
Background knowledge can enter through orientation, probabilistic, or
logical constraints \citep{meek2013causal,claassen2012bayesian,
hyttinen2014constraint}. We study the consequences of treating such
information as fallible.

\paragraph{LLMs and imperfect experts.}
LLMs provide pairwise causal judgments, orderings, and structural priors
\citep{kiciman2023causal,ban2023query,vashishtha2023causal,long2023causal}.
MosaCD filters proposals and propagates accepted orientations
\citep{lyu2025mosacd}. Guess2Graph uses expert graph guesses to guide
statistical tests \citep{hiremath2025guess2graph}. Learning to defer
selects between numerical evidence and an imperfect expert
\citep{clivio2025learning}. Our study asks three separate questions:
whether semantic directions beat a mechanical control, whether the chosen
query locations beat matched alternatives, and whether retained models
support downstream inference. We additionally evaluate Guess2Graph on its compatible native
PC-family skeleton task. This comparison is not matched in semantic-query
format or budget, so we keep it separate from the directional recovery
comparisons.

\paragraph{Calibration and graph uncertainty.}
Learn then Test and conformal risk control make guarantees under explicit
sampling and loss assumptions
\citep{angelopoulos2021learn,angelopoulos2024conformal}.
We use finite-grid concentration for admission loss and a separate rank
calibration for disagreement counts. IDA and methods for partially oriented
graphs propagate graph uncertainty to causal effects
\citep{maathuis2009estimating,perkovic2019mpdag}.
Our retained set allows errors in semantic constraints within a stored
reference. Its guarantee exposes reference exclusion, semantic exclusion,
and parameter uncertainty as distinct failure events.

\section{Methodology}
\label{sec:method}

Let data $D$ follow a distribution $P$ generated by a DAG $G^\star$.
We assume causal sufficiency, the Markov condition, and faithfulness for
the observational identification statements. The experiments use
linear-Gaussian structural equations with unrestricted noise variances.
The empirical recovery policy, admission certificate, and downstream
diagnostics are evaluated separately. They are not one experimentally
validated, certified end-to-end procedure.

\subsection{Query Allocation}
\label{sec:allocation}

Identification is relative to a model class $\mathcal M$. Define
\begin{equation}
\Theta_{\mathcal M}(P)=\{\theta\in\mathcal M:P_\theta=P\}.
\label{eq:identified-class}
\end{equation}
A direction is identified if it is constant on this class.
F1 contains directions that vary across the class. Under our unrestricted
linear-Gaussian model, these are reversible population-CPDAG edges.
F2 contains population-identified directions that an estimator recovers
unreliably at finite sample sizes. A stable fitted direction can be
unidentified. An unstable fitted direction can be identified.

Observational equivalence supplies the motivation for semantic input.
Two faithful Gaussian SCMs can induce the same distribution and orient a
reversible edge oppositely. An observational rule has worst-case error at
least $1/2$ on that paired problem. This is not an accuracy ceiling on
an arbitrary benchmark distribution. Semantic information can distinguish
the worlds or change their prior weights. Appendix~\ref{app:proofs}
states the argument and its assumptions.

Neither population region is observed at deployment. We use undirected
edges of the fitted graph $\widehat{\mathcal C}$ as $\widehat F_1$.
To construct $\widehat F_2$, refit the estimator on $B$ row-bootstrap
samples. For a fitted skeleton edge, count its two directed appearances
as $n_{i\to j},n_{j\to i}$ and its undirected appearances as $n_u$.
Missing adjacencies contribute to none of these counts. Define
\begin{equation}
q_e=\frac{\max\{n_{i\to j},n_{j\to i}\}}
{n_{i\to j}+n_{j\to i}},\qquad
a_e=\frac{n_{i\to j}+n_{j\to i}+n_u}{B}.
\label{eq:locator}
\end{equation}
A fitted committed arrow enters $\widehat F_2$ when
$n_{i\to j}+n_{j\to i}>n_u$ and $q_e<\tau$.
The first condition makes the denominator positive. The primary setting
uses $B=25$ and $\tau=.75$. A sensitivity analysis adds $a_e\ge m_{\rm adj}$.
Neither score is a calibrated probability of orientation error.

F1-only queries fitted unresolved edges. F1+F2 adds selected unstable
committed arrows. F1+rand instead adds randomly selected committed arrows
under the matched realized response-request budget. F1+F2-data uses the
same F2 targets but resolves them from bootstrap directional counts.
Thus the random contrast tests allocation, while the data contrast tests
resolution conditional on the selected targets.

Before adding semantic constraints, relax each selected committed arrow.
Release collider restrictions that would force that arrow and retain the
remaining structural constraints. Removing constraints expands the ideal
completion class. The actual stored reference, denoted
$\widehat\Theta_D$, can still exclude truth through skeleton error,
retained orientations, propagation choices, or enumeration limits.
All retention statements concern this stored reference.

\subsection{Semantic Proposals and Admission}
\label{sec:admission}

Each queried pair receives $R$ responses with balanced display order.
Let $v_+(e)$ and $v_-(e)$ count valid directional responses. The majority
direction is $d(e)$, with score
\begin{equation}
s_e=\frac{\max\{v_+(e),v_-(e)\}}{R}.
\label{eq:agreement}
\end{equation}
UNKNOWN and invalid responses are no-votes. Directional ties and pairs
without a qualifying majority cause abstention. The primary recovery
policy uses $R=10$ and requires at least six identical directional votes.
This agreement rule is empirical. It is not the certified selector below.

The primary policy processes eligible proposals in canonical unordered-pair
order. It accepts a proposal only if the accumulated constraints admit a
valid DAG completion. A timed-out completion check returns unknown and
does not certify admission. Direct admissions form $A'$. Propagated
orientations are recorded separately and do not count as extra semantic
observations. These checks preserve feasibility when started from a
nonempty completion class. They do not establish truth or complete
representation of every permitted DAG.

The full policy must be fixed before calibration. This includes the
query locator, parser, response budget, score, proposal order, and
structural filter. In a downstream diagnostic, the scored orientation
list may instead contain majority-vote proposals without sequential
structural admission. We identify this distinction explicitly in
Section~\ref{sec:downstream-results}; its calibration must use the same
scoring policy as its evaluation.

\subsection{Admission Certification}
\label{sec:certification}

Fix a finite threshold grid $\mathcal T$. On graph task $g$, let $M_g(t)$
count direct admissions at threshold $t$. Let $V_g(t)$ count admissions
that are absent from, or opposite to, the benchmark DAG. Define
\begin{equation}
L_g(t)=\frac{V_g(t)}{\max\{1,M_g(t)\}},\qquad
\widehat R(t)=\frac1n\sum_{g=1}^{n}L_g(t).
\label{eq:admission-loss}
\end{equation}
The loss is zero under complete abstention. Assume that calibration tasks
are independent and follow the target task distribution. Dependence
within a graph task is allowed.

For $\eta\in(0,1)$, the finite-grid upper bound is
\begin{equation}
U_H(t)=\min\!\left\{1,\widehat R(t)+
\sqrt{\frac{\log(|\mathcal T|/\eta)}{2n}}\right\}.
\label{eq:admission-bound}
\end{equation}
Among thresholds satisfying $U_H(t)\le\alpha$, choose the one with the
largest mean task-level admission fraction. Use a fixed tie rule.
If no threshold qualifies, abstain. Hoeffding's inequality and a union
bound give, with calibration probability at least $1-\eta$,
$\mathbb E[L_{\rm new}(\widehat t)]\le\alpha$.
Appendix~\ref{app:proofs} gives the proof. The guarantee controls expected
task loss. It does not promise zero errors on each graph.

Even at zero empirical loss, this certificate requires
\begin{equation}
n\ge\left\lceil\frac{\log(|\mathcal T|/\eta)}{2\alpha^2}\right\rceil.
\label{eq:hoeffding-floor}
\end{equation}
At positive loss, replace $\alpha$ in the denominator by
$\alpha-\widehat R(t)$ when that difference is positive.
This is a certificate-specific requirement. For comparison, an exact
binomial bound on the event $L_g(t)>0$ gives a zero-event bound
$1-(\eta/|\mathcal T|)^{1/n}$. Since
$L_g(t)\le\mathbf1\{L_g(t)>0\}$, that event bound also controls expected
loss. The comparison does not certify any nontrivial policy on our real
pools. Empirical Agreement and Add-one selectors remain separate.

\subsection{Error-Tolerant Retention}
\label{sec:retention}

Hard commitment can fail despite low admission loss. One wrong constraint
on every task with at least $\lceil1/\alpha\rceil$ admissions gives
$L_g\le\alpha$, yet excludes the truth from the zero-disagreement set.
Moreover, an orientation-error count alone does not bound causal-effect
error without restrictions on coefficients or paths. Appendix~\ref{app:proofs}
gives both counterexamples.

Freeze the scoring policy. Use a separate calibration set whose true-DAG
disagreement counts $V_1,\ldots,V_m$ are exchangeable with the target
count, conditional on that policy. For $\delta_S\in(0,1)$, set
\begin{equation}
r=\lceil(m+1)(1-\delta_S)\rceil,\qquad
k=\begin{cases}V_{(r)},&r\le m,\\ \infty,&r>m.\end{cases}
\label{eq:rank}
\end{equation}
For the target orientation list $A'$, define
\begin{equation}
\begin{split}
D_{A'}(G)&=\sum_{e\in A'}\mathbf1\{d_G(e)\ne d(e)\},\\
\Theta^{(k)}&=\{G\in\widehat\Theta_D:D_{A'}(G)\le k\}.
\end{split}
\label{eq:kset}
\end{equation}
An absent edge counts as a disagreement. The choice $k=0$ enforces all
scored constraints; it need not select one DAG. If $k\ge|A'|$, the full
stored reference remains. We call this the \emph{true-DAG disagreement
$k$-set} when calibration uses disagreement with truth.

\begin{theorem}[Reference adequacy and downstream validity]
\label{thm:retention}
Under the exchangeability assumption above, let
$A=\{G^\star\in\widehat\Theta_D\}$ and $p_D=\Pr(A)$.
Then
\begin{equation}
\begin{split}
\Pr(G^\star\in\Theta^{(k)})
&=p_D-\Pr(A\cap\{V_{\rm new}>k\}),\\
\max\{0,p_D-\delta_S\}
&\le\Pr(G^\star\in\Theta^{(k)})\le p_D.
\end{split}
\label{eq:adequacy-bound}
\end{equation}
For a fixed target effect $\theta^\star$, suppose the true-graph member
interval satisfies $\Pr(\theta^\star\in C_{G^\star})\ge1-\delta_P$.
If the reported envelope includes every retained member interval, then
\begin{equation}
\Pr\{\theta^\star\in I(\Theta^{(k)})\}
\ge\max\{0,p_D-\delta_S-\delta_P\}.
\label{eq:effect-bound}
\end{equation}
No independence between these failure events is required.
\end{theorem}

The full proof is given in Appendix~\ref{app:proofs}.
It combines a standard rank bound with set inclusion.
If $p_D\ge1-\delta_D$, the usual retention bound is
$1-\delta_D-\delta_S$. Equation~\eqref{eq:adequacy-bound} also exposes
the ceiling: pruning cannot increase exact truth membership above $p_D$.
The theorem is a marginal guarantee. Without conditional calibration,
it does not assert $1-\delta_S$ survival conditional on $A$.

For nonempty retained sets, the effect envelope is
\begin{equation}
I(\Theta^{(k)})=
\left[\min_{G\in\Theta^{(k)}}\ell_G,
      \max_{G\in\Theta^{(k)}}u_G\right],
\qquad C_G=[\ell_G,u_G].
\label{eq:envelope}
\end{equation}
Unavailable envelopes cause abstention. For an available $[\ell,u]$,
act positively when $\ell>0$ and negatively when $u<0$.
Otherwise abstain. The parameter-validity event is an additional
assumption; bootstrap intervals do not prove it. A guarantee for an
adaptively selected target or action requires the corresponding
simultaneous parameter event.

\section{Experiments}
\label{sec:experiments}

The primary study uses Asia, Sachs, Child, Insurance, and Alarm, five
paired seeds, PC and BOSS, and $N\in\{500,2000\}$. Policies share the
numerical data and base fit within each instance. Usable counts are
25, 20, 25, and 25 for PC-500, PC-2000, BOSS-500, and BOSS-2000.

Primary recovery estimates condition on the usable-instance cohort.
Within each setting, all allocation policies use the same cohort.

Primary semantic acquisition uses Claude Opus~4.5 at temperature $.8$.
Additional-family acquisition uses temperature $0$.

We distinguish pooled instance means from equal-network contrasts.
The latter first average paired effects within a network, then weight
networks equally. Reported 95\% network intervals are exploratory and
are not adjusted for multiple comparisons. Five primary networks and
three additional families limit cross-network inference. Repeated
responses, seeds, and resampling appearances do not create independent
networks. Appendix~\ref{app:implementation} specifies the protocols.

\begin{figure*}[!t]
\centering
\includegraphics[width=\textwidth]{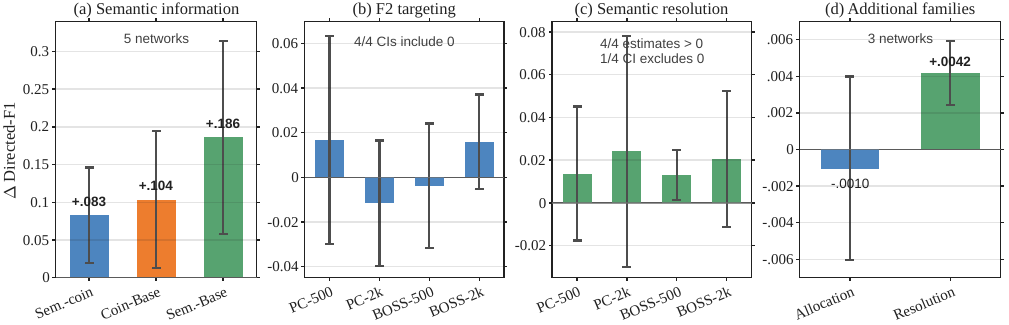}
\caption{\textbf{Semantic information, allocation, and resolution.}
(a) Semantic-minus-coin, coin-minus-Base, and semantic-minus-Base
F1-only contrasts.
(b) F1+F2 minus matched random in the four primary settings.
(c) F1+F2 minus bootstrap-only F2 resolution in the four primary settings.
(d) Allocation and resolution contrasts on the additional families.
Bars show equal-network effects; whiskers show exploratory 95\% network
intervals. Primary contrasts use five networks; additional-family
contrasts use three. Panels use different scales. All displayed
allocation intervals include zero.}
\label{fig:core}
\end{figure*}

\subsection{Semantic Information Improves Orientation}
\label{sec:orientation-results}

The coin-flip control asks whether semantic content adds value beyond
assigning directions to the same F1 candidates.
Semantic minus coin yields $.08263$ Directed-F1
$[.01917,.14609]$ across five networks (Figure~\ref{fig:core}a).
Coin minus Base is also positive. Thus the Base comparison alone would
overstate the information supplied by semantics. The matched candidate
contrast supports an additional semantic benefit within this evaluation;
it is not evidence of generalization beyond these benchmark families.

The native PC-Guess comparison evaluates skeleton recovery
on 35 paired usable instances: 19 at $N=500$ and 16 at
$N=2000$. The observed pooled Skeleton-F1 differences
from the shared PC baseline are $+.0254$ and $+.0142$,
respectively. These descriptive comparisons do not match
our directional task or semantic-query budget.
Appendix~\ref{app:guess2graph},
Table~\ref{tab:guess2graph}, reports absolute metrics,
paired differences, and output availability.

Table~\ref{tab:recovery} places the primary recovery results together
with two simple internal alternatives: a bootstrap--LLM agreement gate
and propagation-based query allocation. Neither alternative uniformly
dominates F1+F2. Full Directed-F1/SHD results, paired contrasts, network
heterogeneity, and budget analyses appear in
Appendix~\ref{app:recovery}.

\begin{table}[t]
\centering\small
\caption{\textbf{Primary graph recovery and internal baselines.}
Pooled Directed-F1 point estimates; usable counts are
25/20/25/25 in column order. The table provides descriptive
recovery comparisons. Exploratory 95\% intervals for the primary
paired, equal-network contrasts appear in Figure~\ref{fig:core}.
These contrasts use different weighting from the pooled means.}
\label{tab:recovery}
\setlength{\tabcolsep}{3pt}
\begin{tabular}{@{}lrrrr@{}}
\toprule
&\multicolumn{2}{c}{PC}&\multicolumn{2}{c}{BOSS}\\
\cmidrule(lr){2-3}\cmidrule(lr){4-5}
Policy&500&2000&500&2000\\
\midrule
Base&.4011&.4962&.6256&.6500\\
F1-only&.6177&.6831&.7744&.8414\\
F1+rand&.6176&\best{.7179}&\best{.7831}&.8450\\
F1+F2-data&.6206&.6837&.7662&.8405\\
Bootstrap--LLM&\second{.6215}&.6867&.7685&\second{.8488}\\
Propagation&.6184&\second{.7064}&\second{.7811}&.8462\\
F1+F2&\best{.6343}&.7031&.7793&\best{.8610}\\
\bottomrule
\end{tabular}
\end{table}

We report pooled point estimates for descriptive comparison.
Uncertainty analysis focuses on paired mechanism contrasts across
networks rather than individual policy means. The exploratory
intervals reflect variation across five benchmark networks and
do not establish generalization beyond them.

\subsection{F2 Targeting Has No Established Advantage}
\label{sec:allocation-results}

We compare F1+F2 with random committed-edge targets under the same
realized response-request budget. All four primary equal-network
intervals include zero, and the point estimates change sign
(Figure~\ref{fig:core}b). The additional-family estimate is
$-.00104$ $[-.00605,.00398]$. Thus, the tested F2 locator does not
establish an advantage over matched-random targeting. This result does
not imply that all instability-based allocation rules are ineffective.

\begin{figure}[t]
\centering
\includegraphics[width=\columnwidth]{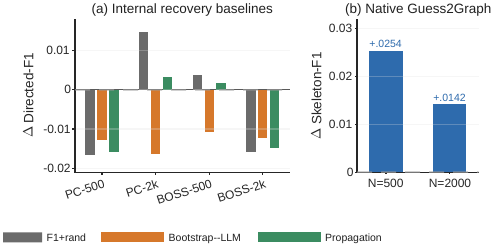}
\caption{\textbf{Baseline comparisons.}
(a) Pooled Directed-F1 changes relative to F1+F2 for three internal
alternatives across the four primary settings.
(b) Native Guess2Graph Skeleton-F1 improvement over its compatible
PC baseline; this is not a matched directional comparison.
Propagation is a frozen-response replay with archived responses
available for 260/285 selected edges.
Both panels report descriptive point estimates without uncertainty
intervals; their metrics and comparison populations differ.}
\label{fig:baseline-ablation}
\end{figure}

We evaluate propagation-based allocation through a frozen-response
replay. It ranks candidate edges by the additional orientations that
their resolution can induce under the frozen propagation rules.
Archived responses are available for 260 of 285 selected edges.
Its recovery point estimates exceed F1+F2 in two settings and fall
below it in two (Table~\ref{tab:recovery}).
Because acquisition is incomplete, this replay does not isolate
locator quality under a fully matched query protocol.
Figure~\ref{fig:baseline-ablation}a summarizes this comparison, and
Appendix~\ref{app:recovery} reports the corresponding diagnostics.

The budget sweep in Appendix~\ref{app:recovery}
(Figure~\ref{fig:budget}) shows no consistent allocation advantage
across response caps. Full allocation is a separate endpoint because
candidate counts vary. Appendix~\ref{app:robustness} reports
locator-threshold, adjacency-support, bootstrap-count, and
proposal-order sensitivity. Its bootstrap-count analysis measures
candidate overlap, not recovery at larger bootstrap counts.

\subsection{Semantic Resolution Can Improve Selected Directions}
\label{sec:resolution-results}

We hold the F2 target set fixed and change its directional evidence.
Semantic resolution has a positive point estimate in all four primary
settings, but only BOSS-500 has an interval above zero
(Figure~\ref{fig:core}c). The separate Barley, Hailfinder, and Win95pts
replay gives $.00418$ $[.00243,.00593]$ over bootstrap-only resolution.
This is a small positive result for the tested additional families.
These families are not globally held out, and their acquisition
temperature differs from the primary study.

We also test a bootstrap--LLM agreement baseline.
It accepts a semantic proposal only when it agrees with the
bootstrap majority and passes the same structural check.
Of 285 eligible F2 proposals, 141 satisfy the agreement rule
and 132 are admitted. Its pooled Directed-F1 is lower than
F1+F2 in all four primary settings (Table~\ref{tab:recovery}).
This recovery comparison does not establish whether the
agreement gate better controls admission risk.

These results separate directional evidence from query location.
Semantic evidence can improve selected directions, but it does not
dominate every statistical alternative. Appendix~\ref{app:recovery}
reports the full semantic-versus-data contrasts, SHD results, network
heterogeneity, and additional-family sensitivity. Appendix~\ref{app:robustness}
reports proposal-order and completion-check sensitivity.

\subsection{Meaningful Variable Content Supports the Signal}
\label{sec:semantic-results}

\begin{figure*}[!t]
\centering
\includegraphics[width=\textwidth]{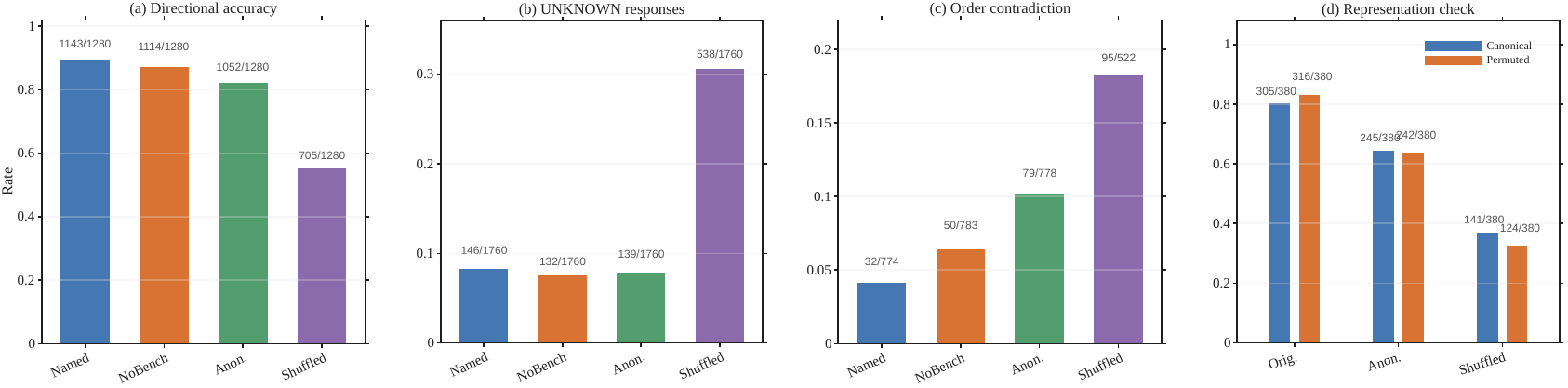}
\caption{\textbf{Semantic content and representation controls.}
(a) Raw directional accuracy uses 1,280 truth-adjacent requests per
primary condition. (b) UNKNOWN uses all 1,760 completed requests per
condition. (c) Order contradiction uses only pairs with two parsed
directions; denominators are printed at each point. (d) Additional-family
raw accuracy uses 380 requests per condition and representation.
Primary temperature is $.8$; additional-family temperature is $0$.
Values are pooled descriptive proportions, without independent-request
error bars. These panels do not share a statistical denominator.}
\label{fig:semantics}
\end{figure*}

We vary benchmark identity, variable identity, and description correspondence.
Named inputs outperform No-Benchmark, Anonymous, and Shuffled inputs in
raw directional accuracy (Figure~\ref{fig:semantics}a).
The Named--Shuffled degradation is much larger than the
Named--No-Benchmark degradation. UNKNOWN and presentation contradiction
measure different response behaviors and use different denominators
(Figure~\ref{fig:semantics}b--c). This pattern supports useful variable
content, but does not rule out benchmark familiarity or ordering cues.

The additional-family representation check preserves the advantage of
Original descriptions under canonical and permuted representations
(Figure~\ref{fig:semantics}d). Exact representation invariance does not
hold. These controls test sensitivity to content and representation;
they are not a clean test of whether benchmark information appeared in
model training. Appendix~\ref{app:semantic} gives the counts and
recovery-level perturbation results.

\subsection{Admission Certification Depends on Independent Tasks}
\label{sec:admission-results}

The real calibration pools contain ten F1 and nine F2 graph tasks.
Neither pool meets the zero-loss requirement of the tested Hoeffding
certificate at the reported risk levels (Table~\ref{tab:admission}).
The Hoeffding-certified selector therefore abstains. Positive observed loss increases
the required task count further. The bound concerns independent
target-relevant graph tasks, not the number of queried edges or responses.

The zero-event comparison in Table~\ref{tab:admission} shows why these
requirements should not be presented as universal limits. The tighter
bound still needs more tasks than either real pool, even at zero events.
Empirical Agreement and Add-one policies admit proposals with nonzero
observed loss, but do not inherit the tested certificate.
Appendix~\ref{app:calibration} separates those descriptive results from
the formal guarantee.

\begin{table}[t]
\centering\small
\caption{\textbf{Calibration feasibility.}
Independent-task requirements with six fixed thresholds and
$\eta=.1$. The exact zero-event row bounds the event of any task loss;
it is a theoretical comparison, not a certified real-data result.
Positive empirical loss raises the Hoeffding requirement.}
\label{tab:admission}
\setlength{\tabcolsep}{4pt}
\begin{tabular}{@{}lrrr@{}}
\toprule
Target risk $\alpha$&.10&.20&.30\\
\midrule
\theory{Hoeffding, zero loss}
&\theory{205}
&\theory{52}
&\theory{23}\\
\theory{Exact, zero loss events}
&\theory{39}
&\theory{19}
&\theory{12}\\
\midrule
\empirical{F1 available}
&\multicolumn{3}{c}{\empirical{10 graph tasks}}\\
\empirical{F2 available}
&\multicolumn{3}{c}{\empirical{9 graph tasks}}\\
\abstain{Hoeffding selector}
&\multicolumn{3}{c}{\abstain{Complete abstention}}\\
\bottomrule
\end{tabular}
\end{table}

\subsection{Reference Adequacy Limits Exact Truth Retention}
\label{sec:downstream-results}

\paragraph{Fresh Oracle retention.}
The Oracle reference is the exact equivalence class of the true DAG.
We evaluate 240 draws on Asia, Sachs, and Child, with a fresh $R=15$
response bank for each draw. All 38,400 planned response slots
completed. This evaluator scores majority-vote proposals without the
primary sequential structural-admission procedure. It is a retention
diagnostic, not an end-to-end pipeline result.

Under the truth-containing Oracle reference, calibrated retention
preserves truth in every evaluated draw. Effect coverage is .9831
for both Data-PI and the calibrated set, while conditional width is
1.2633 and 1.1965, respectively
(Table~\ref{tab:oracle-full}). The calibrated set leaves Asia and Child unchanged relative to
Data-PI; its pruning, width, and action changes occur in Sachs. All 150 calibration splits select
$k=4$. The fixed-$k$ sweep reaches full truth retention at $k=4$
while retaining fewer graphs than the full reference
(Table~\ref{tab:oracle-k-sweep}). These results are descriptive and
use three fixed networks. Appendix~\ref{app:oracle} gives the full
protocol and per-network analysis.

\paragraph{Learned references.}
The learned X6 study restricts references to exact MEC extensions and
does not perform F2 relaxation. Of 80 planned tasks, 72 are downstream
usable. Truth is present in $3/80=.0375$ planned tasks, or
$3/72=.0417$ usable tasks. Its reference-disagreement $k$-set retains
the closest member with frequency $.9965$, yet effect coverage remains
$.6028$. The closest member is selected using benchmark truth, so
closest-member retention is a proxy diagnostic rather than
evidence of exact truth retention.

The learned-reference fixed-$k$ sweep exposes a different limitation.
Increasing $k$ improves closest-member retention, but it cannot restore
a true DAG that is absent from the stored reference. Appendix~\ref{app:x6}
reports the full X6 downstream table, tolerance sweep, retained-set
sizes, availability analysis, and reference-construction ablations.

The separate true-DAG disagreement diagnostic has marginal
post-retention truth coverage $.0396$ over evaluation appearances and
conditional survival $1.0000$, but only \emph{three distinct tasks}
contain truth upstream. Thus, retention can preserve available truth;
it cannot repair reference exclusion.

Broader constructions raise planned exact-truth coverage from
$.0375$ to at most $.0750$, while mean stored reference size rises
from $44.6$ to $4291.4$. Appendix~\ref{app:x6} gives the full
construction comparison.

\begin{figure}[t]
\centering
\includegraphics[width=\columnwidth]{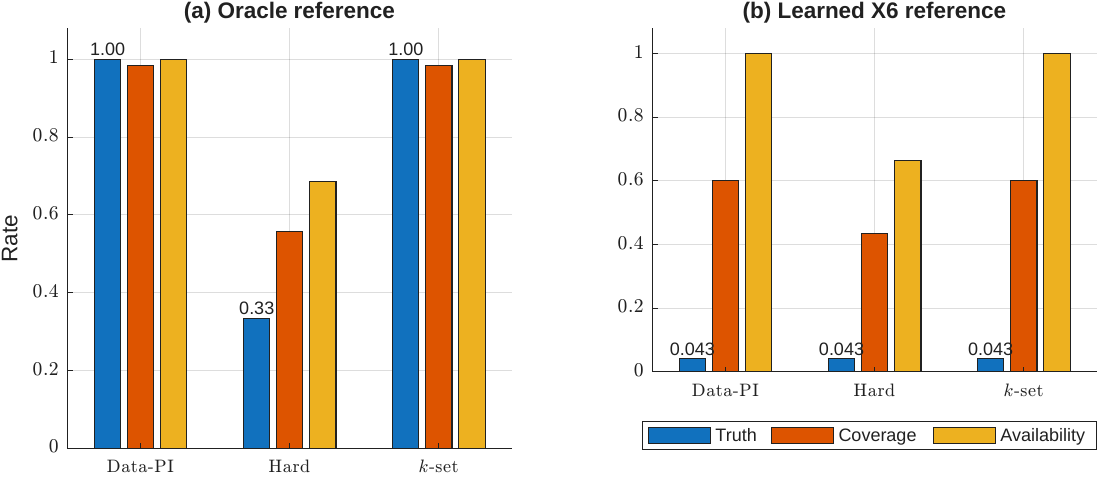}
\caption{\textbf{Retention depends on reference adequacy.}
Under the Oracle reference, calibrated tolerance restores exact truth
retention while preserving effect coverage. Under the learned X6
reference, exact truth retention remains low because the stored
reference usually excludes the true DAG. The panels use different
evaluation populations and are descriptive.}
\label{fig:retention-reference}
\end{figure}

Figure~\ref{fig:retention-reference} summarizes the contrast:
tolerance restores exact truth retention when truth is available in
the Oracle reference, but cannot recover truth that is absent from the
learned X6 reference.

\section{Discussion and Conclusion}
\label{sec:conclusion}

Semantic orientation improves recovery over the coin-flip control, but
the tested F2 locator does not establish an advantage over
matched-random allocation. The agreement and propagation comparisons
are diagnostics of specific policies rather than general rankings.
Admission certification requires assumptions and numbers of independent
tasks that the current calibration pools do not establish. Downstream
retention also depends on reference adequacy: pruning can preserve an
available true graph but cannot recover one that is absent from the
stored reference.

These results support semantic assistance as a component of causal
discovery rather than as an unrestricted replacement for statistical
evidence. Potential applications include prioritizing ambiguous causal
directions, combining domain knowledge with observational discovery,
and maintaining uncertainty sets for downstream effect estimation and
decision support. Future work should study learned or decision-aware
query allocation, stronger reference construction, and admission rules
calibrated on substantially larger collections of independent graph
tasks. Evaluation should also extend to more causal domains and less
benchmark-familiar semantic models. The small number of benchmark
families, possible semantic familiarity, and incomplete primary
protocol records limit broader claims.

\FloatBarrier
\section*{AI Use Statement}
Generative AI tools assisted software development and manuscript preparation.
Experimental LLM judgments are identified separately in the methods and
reproducibility sections. The authors are responsible for the final
methodology, numerical results, references, and scientific claims.

\clearpage
\bibliography{references}
\clearpage
\section*{Checklist}

\begin{enumerate}

\item For all models and algorithms presented, check if you include:
\begin{enumerate}
  \item A clear description of the mathematical setting, assumptions,
  algorithm, and/or model. \textbf{[Yes]}
  
  The mathematical setting, assumptions, semantic-query policies,
  admission rules, and retention procedures are described in
  Sections~\ref{sec:method} and the corresponding appendices.

  \item An analysis of the properties and complexity (time, space,
  sample size) of any algorithm. \textbf{[Yes]}
  
  The paper analyzes task-count requirements for admission certification,
  retained-set sizes, enumeration limits, and empirical computational
  constraints. Relevant details appear in
  Section~\ref{sec:admission-results} and the appendices.

  \item (Optional) Anonymized source code, with specification of all
  dependencies, including external libraries. \textbf{[No]}
  
  The manuscript reports implementation versions and dependencies, but
  the retained submission record does not contain a complete anonymized
  release of all code required for exact reproduction. We commit to open source code after the acceptance of the paper.
\end{enumerate}

\item For any theoretical claim, check if you include:
\begin{enumerate}
  \item Statements of the full set of assumptions of all theoretical
  results. \textbf{[Yes]}
  
  Identification, task-independence, exchangeability, reference, and
  parameter assumptions are stated in Section~\ref{sec:method}.

  \item Complete proofs of all theoretical results. \textbf{[Yes]}
  
  Complete proofs and counterexamples used by the paper are provided in
  Appendix~\ref{app:proofs}.

  \item Clear explanations of any assumptions. \textbf{[Yes]}
  
  The assumptions and their scope are discussed in the main method
  section and the associated appendices.
\end{enumerate}

\item For all figures and tables that present empirical results, check if
you include:
\begin{enumerate}
  \item The code, data, and instructions needed to reproduce the main
  experimental results (either in the supplemental material or as a URL).
  \textbf{[No]}
  
  The manuscript provides substantial implementation and protocol detail,
  but the retained record does not contain every primary task-level file,
  the coin-control replicate count and randomization seeds, or every native
  baseline artifact. We will open source code after acceptance of the paper.

  \item All the training details (e.g., data splits, hyperparameters, how
  they were chosen). \textbf{[Yes]}
  
  Data-generation settings, discovery hyperparameters, bootstrap settings,
  semantic-response settings, calibration splits, and evaluation protocols
  are reported in Appendix~\ref{app:implementation} and the experiment-specific
  appendices.

  \item A clear definition of the specific measure or statistics and error
  bars (e.g., with respect to the random seed after running experiments
  multiple times). \textbf{[Yes]}
  
  Directed-F1, SHD, weighting conventions, conditional width, coverage,
  and exploratory network intervals are defined in
  Appendix~\ref{app:implementation}. Figures without supported uncertainty
  intervals are identified as descriptive.

  \item A description of the computing infrastructure used.
  (e.g., type of GPUs, internal cluster, or cloud provider).
  \textbf{[Yes]}
  
  Appendix~\ref{app:implementation} reports the available implementation
  record, including Python version, ARM64 macOS environment, library
  versions, model provider, and model settings.
\end{enumerate}

\item If you are using existing assets (e.g., code, data, models) or
curating/releasing new assets, check if you include:
\begin{enumerate}
  \item Citations of the creator If your work uses existing assets.
  \textbf{[Yes]}
  
  Existing causal-discovery methods, benchmark networks, language models,
  and external baselines are cited in the manuscript.

  \item The license information of the assets, if applicable.
  \textbf{[No]}
  
  License information is not reported uniformly for all external assets.

  \item New assets either in the supplemental material or as a URL, if
  applicable. \textbf{[Not Applicable]}
  
  The paper does not claim to release a new dataset or benchmark asset.

  \item Information about consent from data providers/curators.
  \textbf{[Not Applicable]}
  
  The main study uses benchmark graph structures and simulated
  observational samples rather than newly collected human data.

  \item Discussion of sensible content if applicable, e.g., personally
  identifiable information or offensive content.
  \textbf{[Not Applicable]}
  
  The evaluated benchmark structures and simulated samples do not contain
  personally identifiable or participant-generated sensitive content.
\end{enumerate}

\item If you used crowdsourcing or conducted research with human subjects,
check if you include:
\begin{enumerate}
  \item The full text of instructions given to participants and screenshots.
  \textbf{[Not Applicable]}

  \item Descriptions of potential participant risks, with links to
  Institutional Review Board (IRB) approvals if applicable.
  \textbf{[Not Applicable]}

  \item The estimated hourly wage paid to participants and the total amount
  spent on participant compensation.
  \textbf{[Not Applicable]}
\end{enumerate}

\end{enumerate}

\appendix

% A. PROOFS AND SUPPORTING THEORY
% ============================================================
\clearpage
\appendix

\section{Deferred Proofs}
\label{app:proofs}

\subsection{Reference Adequacy and Retention}
\begin{proof}[Proof of Theorem~\ref{thm:retention}]
Condition on the frozen scoring policy. Exchangeability of
$V_1,\ldots,V_m,V_{\rm new}$ gives the conservative rank bound
$\Pr(V_{\rm new}>k)\le\delta_S$. Ties can only make the non-strict
retention event more conservative. If $r=m+1$, the definition
$k=\infty$ makes the statement immediate.

On $A=\{G^\star\in\widehat\Theta_D\}$, the true graph belongs to
$\Theta^{(k)}$ exactly when $V_{\rm new}\le k$. Outside $A$, it cannot
belong to a subset of $\widehat\Theta_D$. Therefore
\[
\{G^\star\in\Theta^{(k)}\}=A\cap\{V_{\rm new}\le k\}.
\]
Taking probabilities gives the identity in
Equation~\eqref{eq:adequacy-bound}. The lower bound follows from
$\Pr(A\cap\{V_{\rm new}>k\})\le\delta_S$. The upper bound follows
from $\Theta^{(k)}\subseteq\widehat\Theta_D$.

Let $E_P=\{\theta^\star\in C_{G^\star}\}$. On truth retention and
$E_P$, the envelope includes $C_{G^\star}$ and hence $\theta^\star$.
A union bound gives Equation~\eqref{eq:effect-bound}. No independence
between reference selection, disagreement, and parameter events is used.
Averaging over a policy frozen on separate data preserves the claim when
the conditional assumptions hold.
\end{proof}

For $p_D>0$, the same argument gives
\[
\Pr(G^\star\in\Theta^{(k)}\mid A)
\ge\max\{0,1-\delta_S/p_D\}.
\]
It does not generally give $1-\delta_S$ after conditioning on reference
membership. That stronger statement needs calibration or exchangeability
appropriate to the conditional population. Empirical conditional survival
in Section~\ref{sec:downstream-results} is not a substitute for this
assumption. Reference coverage bounds exact graph retention from above;
it need not bound empirical effect coverage from above. A wrong graph can
still produce an interval that contains the target effect.

\subsection{Observational Indistinguishability}
\begin{proposition}[Paired observational indistinguishability]
\label{prop:indistinguishability}
Under faithful linear-Gaussian SCMs with independent noise and unrestricted
noise variances, a reversible CPDAG edge admits two DAG parameterizations
with the same observational distribution and opposite directions. The
minimax directional error of an observational binary rule on this pair
is $1/2$.
\end{proposition}
\begin{proof}
Choose two Markov-equivalent DAGs that orient the edge oppositely.
A positive Gaussian distribution faithful to their equivalence class
admits recursive linear parameterizations for both DAGs. Thus the
observational input to the rule has the same distribution in both worlds.
If the rule selects one direction with probability $q$, its errors are
$q$ and $1-q$ on the two worlds. Their maximum is at least $1/2$.
A fair coin attains equality.
\end{proof}
This standard argument is a worst-case paired-world statement. It does
not limit accuracy under a particular benchmark prior. Additional
identifying assumptions or informative semantic side information change
the problem. Identical observational and semantic inputs do not.

\subsection{Structural Feasibility}
\begin{proposition}[Feasible sequential admission]
\label{prop:feasibility}
Start with a nonempty completion class. If each accepted update retains
at least one completion and accumulates earlier commitments, the class
remains nonempty and all accepted commitments remain satisfied.
\end{proposition}
\begin{proof}
The initial class is nonempty. The admission condition preserves this
property at every accepted update. Accumulating constraints prevents a
later completion from violating an earlier commitment. Induction gives
the result.
\end{proof}
The proposition concerns the class checked by the procedure. It neither
establishes truth membership nor proves that a propagation shortcut
represents every permitted completion. An independently truncated stored
reference also requires its own membership check.

\paragraph{Relaxation and propagation.}
Let a reference permit $A\to B\leftarrow C$ with nonadjacent $A,C$.
After releasing the collider restriction and erasing $C\to B$, the
relaxed constraints can permit both that collider and $A\to B\to C$.
Applying ordinary CPDAG rule R1 to $A\to B-C$ would force the chain.
That step would remove a permitted collider from this relaxed class.
Thus standard CPDAG propagation and nonempty completion checks do not
alone prove complete representation of a class spanning several MECs.
Our retention theorem concerns the actual stored set, including any
such exclusions.

\subsection{Finite-Grid Admission Risk}
\begin{proposition}[Finite-grid admission certificate]
\label{prop:admission-certificate}
Under independent identically distributed graph tasks and a fixed finite
grid, the selector in Section~\ref{sec:certification} satisfies
$\Pr_{\rm cal}\{\mathbb E[L_{\rm new}(\widehat t)]\le\alpha\}
\ge1-\eta$. Complete abstention is used when no threshold qualifies.
\end{proposition}
\begin{proof}
For a fixed $t$, $L_g(t)\in[0,1]$. Hoeffding's inequality bounds the
probability that the target mean exceeds
$\widehat R(t)+\sqrt{\log(|\mathcal T|/\eta)/(2n)}$
by $\eta/|\mathcal T|$. A union bound makes this event simultaneous
over the grid. Any selected threshold satisfying the bound then has
target risk at most $\alpha$. Maximizing admission among qualifying
thresholds does not change the simultaneous event. If none qualifies,
the abstaining policy has zero loss.
\end{proof}
This proof does not require independent edges or independent responses
within a task. It does require independent calibration tasks with the
specified target distribution. A collection of related named benchmarks
does not establish that assumption by itself.

\paragraph{Exact zero-event comparison.}
Set $Z_g(t)=\mathbf1\{L_g(t)>0\}$ and $\beta=\eta/|\mathcal T|$.
Since $L_g(t)\le Z_g(t)$, a binomial upper bound on
$p_t=\mathbb E[Z_g(t)]$ also bounds expected loss.
For zero events in $n$ independent tasks, the one-sided exact upper
bound is $1-\beta^{1/n}$. A union correction over the grid gives
\[
n\ge\left\lceil\frac{\log\beta}{\log(1-\alpha)}\right\rceil.
\]
With six thresholds and $\eta=.1$, this yields 39, 19, and 12 for
$\alpha=.1,.2,.3$. At nonzero event counts this procedure targets an
event probability, rather than treating fractional graph losses as
Bernoulli observations. Directly applying a binomial model to the
fractional losses would be unjustified.

\subsection{Hard Commitment and Effect Error}
\begin{proposition}[Low admission loss does not imply hard retention]
\label{prop:hard}
If truth is in the reference and at least one scored semantic constraint
disagrees with truth, hard retention excludes truth. For any
$\alpha>0$, a bound $\mathbb E[L]\le\alpha$ alone gives no positive
uniform lower bound on zero-error probability when admission counts
are unrestricted.
\end{proposition}
\begin{proof}
The true DAG fails the zero-disagreement condition when any scored
constraint is wrong. Now consider tasks with exactly one error and
$M\ge\lceil1/\alpha\rceil$ admissions. Their loss is $1/M\le\alpha$
on every task, but their probability of no error is zero.
\end{proof}

\begin{proposition}[Error placement affects downstream effects]
\label{prop:placement}
For every integer $h\ge1$, there are faithful linear-Gaussian examples
with exactly $h$ orientation errors that leave a target effect unchanged,
and examples with exactly $h$ errors that change it. Without coefficient
or path restrictions, the count alone gives no finite uniform bound on
effect error.
\end{proposition}
\begin{proof}
Let $X=\varepsilon_X$ and $Y=bX+\varepsilon_Y$ for independent standard
Gaussian noises and $b\ne0$. The effect of $X$ on $Y$ is $b$.
The same observational distribution admits the reverse representation
\[
Y=U,\qquad X=\frac{b}{1+b^2}Y+W,
\]
where $U\sim\mathcal N(0,1+b^2)$ and
$W\sim\mathcal N(0,(1+b^2)^{-1})$ are independent. In the reverse
DAG, the effect of $X$ on $Y$ is zero. Add $h$ independent two-node
components with nonzero coefficients. Reversing the $h$ added edges
does not change the target effect. Reversing $X\to Y$ and $h-1$ added
edges gives exactly $h$ errors and effect error $|b|$. The nonzero
component models are faithful. Taking $|b|$ arbitrarily large proves
the absence of a count-only uniform bound.
\end{proof}
This is an existence result in the unrestricted model class. It does
not claim arbitrarily large effects in the bounded, standardized
simulation protocol used for primary recovery.

\subsection{Actions under Retained Uncertainty}
\begin{proposition}[Joint action and harm]
\label{prop:action}
Suppose population benefits $b_G(a)$ are known for every retained graph.
Act only when the retained set is nonempty and
$\inf_{G\in\Theta^{(k)}}b_G(a)>0$. Then the joint probability of
acting and causing nonpositive benefit is at most
$1-p_D+\delta_S$, capped at one. If benefit is estimated, add the
failure probability of a parameter event valid simultaneously for the
considered actions.
\end{proposition}
\begin{proof}
On truth retention, $G^\star$ occurs in the infimum. Positivity of
the infimum then implies positive true benefit. The joint adverse
event can occur only when truth retention fails. Apply
Theorem~\ref{thm:retention}. With estimated benefits, the same argument
also requires the simultaneous parameter event.
\end{proof}
This is a bound on acting and causing harm jointly. It is not a bound
on harm conditional on action. Zero observed adverse events do not
establish zero population risk.

% ============================================================
% B. IMPLEMENTATION AND REPRODUCIBILITY
% ============================================================
\section{Implementation and Reproducibility}
\label{app:implementation}

\subsection{Data and Discovery}
Primary recovery uses the structures of Asia, Sachs, Child, Insurance,
and Alarm. Each network has five paired seeds at $N=500,2000$.
Data follow linear-Gaussian structural equations. Noise standard
deviations are $\sqrt{\exp(U)}$ with $U\sim\mathrm{Unif}(-2,2)$.
Parent coefficient magnitudes are drawn from $\mathrm{Unif}(.5,1.2)$
with independent random signs. Variables are standardized before
discovery. Policies share the data and base fit within each instance.

The primary implementation uses \texttt{causal-learn} 0.1.4.8 and
NumPy 1.26.4. PC uses Fisher-$z$ tests at $.05$ and stable-PC.
BOSS uses \texttt{local\_score\_BIC\_from\_cov}. Each estimator fit
has a 60-second timeout. Barley, Hailfinder, and Win95pts supply the
additional-family study. They also occur in calibration diagnostics
and are not claimed to be globally held out or unseen by the LLM.

\begin{table}[t]
\centering\small
\caption{\textbf{Benchmark structures.} Counts describe the benchmark DAGs.}
\label{tab:networks}
\begin{tabular}{@{}lrrl@{}}
\toprule
Network&Nodes&Edges&Recovery panel\\
\midrule
Asia&8&8&Primary\\
Sachs&11&17&Primary\\
Child&20&25&Primary\\
Insurance&27&52&Primary\\
Alarm&37&46&Primary\\
Barley&48&84&Additional\\
Hailfinder&56&66&Additional\\
Win95pts&76&112&Additional\\
\bottomrule
\end{tabular}
\end{table}

\subsection{Locators and Policy Controls}
The F2 bootstrap resamples rows. Missing adjacencies contribute zero to
all directional and undirected appearance counts. The primary settings
are $B=25$ and $\tau=.75$. The additional support requirement is used
only in its sensitivity analysis.

F1+rand permutes the fitted committed-arrow list using an
instance-specific seed and samples without replacement. It selects up
to the number of F2 targets and applies the same relaxation and
admission operations. F1+F2-data proposes $i\to j$ when
$n_{i\to j}\ge n_{j\to i}$ and the reverse otherwise. Its F1
semantic proposals remain unchanged. Canonical unordered-pair ordering
is part of the frozen recovery policy.

The coin control assigns random directions to the F1 candidates.
Replicate outcomes are aggregated within each numerical task before
network aggregation. Its randomized replicates are not independent
graph tasks. The archived summary contains 95 eligible tasks across
five networks, but does not expose the number of coin replicates or
their randomization seeds. The retained record also does not establish whether the coin proposals
used exactly the same proposal-ordering and structural-admission path
as the semantic proposals.

Numerical budgets 10/20/40/80 count planned response requests. At
$R=10$, they permit at most 1/2/4/8 queried edges. Candidate exhaustion
can reduce realized use. Exact realized-request matching was audited
for the primary F1+F2 versus F1+rand comparison in 24 cells. Matching
requests does not match tokens, retries, or provider cost.

\subsection{Responses and Structural Checks}
Primary recovery and semantic controls use Claude Opus~4.5 through
OpenRouter at temperature $.8$. Additional-family acquisition uses
temperature $0$. Primary edges receive ten responses with balanced
presentation order. The primary system prompt is
``Return exactly one requested causal-direction label and no explanation.''
The user prompt supplies the benchmark/domain context and the two variables
and requests exactly one of \texttt{A\_TO\_B}, \texttt{B\_TO\_A}, or
\texttt{UNKNOWN}. A proposal requires six identical valid votes.
UNKNOWN, invalid outputs, and unparseable outputs are no-votes.
The score denominator is the planned response count.

Only direct admissions contribute to the admission loss. Propagated
arrows are separate. A completion timeout returns unknown and does
not certify admission. The X6 diagnostic uses exact-MEC enumeration
rather than unrestricted acyclic skeleton orientations. The learned X6
diagnostic does not apply primary F2 relaxation. The fresh Oracle
diagnostic does not apply primary sequential structural admission.

\subsection{Metrics and Weighting}
Directed-F1 scores directed-edge recovery as
$2\mathrm{TP}/(2\mathrm{TP}+\mathrm{FP}+\mathrm{FN})$.
SHD measures structural discrepancy under the frozen evaluator.
Pooled recovery means weight usable numerical instances equally.
Equal-network contrasts first average paired differences within each
network. The network interval is
\[
\overline\Delta\ \pm\ t_{J-1,.975}\,
\frac{s_{\Delta}}{\sqrt J},
\]
where $J$ is the number of networks. These intervals are exploratory;
the named networks are not a random sample of all causal domains.

Missing or infeasible recovery results remain missing. They are not
zero recovery. Genuine no-op instances remain in paired analyses when
both outcomes are defined. Leave-one-network-out summaries are
influence checks, not new replications.

Let $A_{gs}$ indicate an available interval and let $W_{gs}$ be its
width on task $g$ in evaluation split $s$. Within a population,
conditional mean width is formed from available intervals only.
The historical zero-coded score instead averages $A_{gs}W_{gs}$,
with zero assigned when unavailable. These quantities are reported
separately. Coverage assigns an unavailable interval a miss; action
and joint act-and-harm are zero when the procedure abstains.

Oracle main summaries weight evaluation appearances within a network,
then weight the three networks equally. The Oracle
network-summary file instead averages draw-level summaries equally
within a network. X6 policy summaries weight evaluation appearances
over the 72 usable tasks. These conventions explain small differences
between summaries and must not be interchanged.

\subsection{Reproducibility Record and Scope}
The implementation record identifies source commit
\texttt{bf7b26b7846d471d4bf88ac8c40013313399ea7f}
with an uncommitted correction patch, Python 3.11.15, and an ARM64
macOS environment. The patch is required with the commit.
The reported continuation test suite passes 53 tests; the fresh
Oracle test report passes 58. These are reported implementation
checks, not statistical replications. They cover exact-MEC handling,
truth lookup, conditional-width accounting, response-bank identity,
and missing-value handling.

The available evidence record contains summary tables, the fresh
Oracle draw and response records, and relevant finalizer code. It does
not contain every primary task-level file or the full native baseline
implementation. The primary prompt and implementation settings are recoverable from the
available code record. The available summaries do not expose the coin replicate count,
its randomization seeds, or enough task-level provenance to verify
that its proposal-ordering and structural-admission path exactly
matches the semantic control. The native Guess2Graph version also
cannot be reconstructed from the retained records.

% ============================================================
% C. PRIMARY RECOVERY, ALLOCATION, AND RESOLUTION
% ============================================================
\section{Primary Recovery, Allocation, and Resolution}
\label{app:recovery}

\begin{figure}[t]
\centering
\includegraphics[width=\columnwidth]
{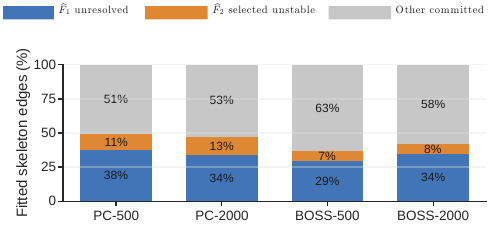}
\caption{\textbf{Fitted-edge composition.}
Fraction of fitted skeleton edges that are unresolved
($\widehat F_1$), selected as bootstrap-unstable
($\widehat F_2$), or other committed edges.
Values are averaged within networks and then
across networks equally.}
\label{fig:edge-composition}
\end{figure}

\begin{table*}[t]
\centering\small
\caption{\textbf{Primary recovery values.}
Each cell is Directed-F1 / SHD. Pooled usable counts are
25/20/25/25 in column order. Bootstrap--LLM and Propagation are
the additional internal baselines reported in the main text.}
\label{tab:primary-full}
\begin{tabular}{@{}lcccc@{}}
\toprule
Policy&PC-500&PC-2000&BOSS-500&BOSS-2000\\
\midrule
Base
&.4011 / 23.44
&.4962 / 20.75
&.6256 / 15.00
&.6500 / 14.88\\

F1-only
&.6177 / 17.60
&.6831 / 15.50
&.7744 / 10.08
&.8414 / 8.24\\

F1+rand
&.6176 / 17.68
&\best{.7179} / \best{14.75}
&\best{.7831} / \second{10.04}
&.8450 / 8.20\\

F1+F2-data
&.6206 / 17.60
&.6837 / 15.55
&.7662 / 10.32
&.8405 / \second{7.92}\\

Bootstrap--LLM
&\second{.6215} / 17.88
&.6867 / 15.75
&.7685 / 10.76
&\second{.8488} / 8.24\\

Propagation
&.6184 / \second{17.36}
&\second{.7064} / \second{14.95}
&\second{.7811} / \best{9.96}
&.8462 / 8.12\\

F1+F2
&\best{.6343} / \best{17.08}
&.7031 / 15.00
&.7793 / \second{10.04}
&\best{.8610} / \best{7.44}\\
\bottomrule
\end{tabular}
\end{table*}

\subsection{Internal Baseline Diagnostics}

Bootstrap--LLM applies a stricter agreement gate to the eligible F2
semantic proposals. Propagation changes only the query-allocation rule
and is evaluated as a frozen-response replay. The latter is not a fully
matched acquisition experiment because archived semantic responses are
unavailable for some selected edges.

\begin{table}[t]
\centering\small
\caption{\textbf{Internal baseline diagnostics.}
Counts summarize the proposal or replay populations used by the two
additional internal baselines. Propagation response availability is
reported separately from its selected query budget.}
\label{tab:baseline-diagnostics}
\setlength{\tabcolsep}{4pt}
\begin{tabular}{@{}lrrr@{}}
\toprule
Baseline & Initial & Intermediate & Final\\
\midrule
Bootstrap--LLM
&285 eligible
&\theory{141 agree}
&\empirical{132 admitted}\\
Propagation
&285 selected
&\empirical{260 responses}
&\abstain{25 missing}\\
\bottomrule
\end{tabular}
\end{table}

The agreement gate therefore retains fewer semantic proposals than the
primary policy. Propagation selects 285 edges, of which 260 have archived
semantic responses and 25 do not. The 25 missing-response edges are still
relaxed from their fitted orientations, but no replacement semantic proposal
is supplied for them. They can be reoriented only through propagation from
other admitted proposals. Thus, the propagation results are a frozen-response
replay with incomplete semantic acquisition, not a fully matched locator
comparison.

\subsection{Primary Paired Contrasts}

\begin{table*}[t]
\centering\small
\caption{\textbf{Primary mechanism contrasts.}
Pooled and equal-network Directed-F1 differ when usable seed counts
differ. Intervals are exploratory 95\% network intervals.}
\label{tab:primary-contrasts}
\setlength{\tabcolsep}{5pt}
\begin{tabular}{@{}llrrrr@{}}
\toprule
Contrast&Setting&Pooled F1&Equal-network F1&95\% interval&Pooled SHD\\
\midrule
F2 minus random
&PC-500&+.01672&+.01672&$[-.02986,.06330]$&$-.60$\\
&PC-2000&$-.01479$&$-.01158$&$[-.03972,.01656]$&+.25\\
&BOSS-500&$-.00379$&$-.00379$&$[-.03174,.02416]$&.00\\
&BOSS-2000&+.01596&+.01596&$[-.00526,.03718]$&$-.76$\\
\midrule
Semantic minus data
&PC-500
&\empirical{+.01375}
&\empirical{+.01375}
&$[-.01759,.04510]$
&\empirical{$-.52$}\\

&PC-2000
&\empirical{+.01940}
&\empirical{+.02406}
&$[-.03006,.07819]$
&\empirical{$-.55$}\\

&BOSS-500
&\best{+.01312}
&\best{+.01312}
&\textcolor{bestblue}{\textbf{$[.00154,.02471]$}}
&\best{$-.28$}\\

&BOSS-2000
&\empirical{+.02054}
&\empirical{+.02054}
&$[-.01146,.05254]$
&\empirical{$-.48$}\\
\bottomrule
\end{tabular}
\end{table*}

\begin{table}[t]
\centering\small
\caption{\textbf{F1 coin-flip control.}
Equal-network Directed-F1 contrasts across 95 eligible tasks and five
networks.}
\label{tab:coin}
\begin{tabular}{@{}lrc@{}}
\toprule
Contrast&Estimate&95\% interval\\
\midrule
\textbf{\textcolor{bestblue}{Semantic minus coin}}
&\best{+.08263}
&\textcolor{bestblue}{\textbf{$[.01917,.14609]$}}\\

\textcolor{secondorange}{Coin minus Base}
&\second{+.10363}
&\textcolor{secondorange}{$[.01250,.19476]$}\\

\textcolor{bestblue}{Semantic minus Base}
&\textcolor{bestblue}{+.18626}
&\textcolor{bestblue}{$[.05838,.31413]$}\\
\bottomrule
\end{tabular}
\end{table}

\subsection{Network Heterogeneity and Influence}

\begin{table*}[t]
\centering\small
\caption{\textbf{Network heterogeneity.}
Mean paired Directed-F1 over available seeds. The upper block tests
allocation; the lower block tests semantic resolution. Entries retain
the precision of the archived table.}
\label{tab:heterogeneity}
\begin{tabular}{@{}lrrrr@{}}
\toprule
Network&PC-500&PC-2000&BOSS-500&BOSS-2000\\
\midrule
\multicolumn{5}{l}{\textbf{F1+F2 minus F1+rand}}\\
Asia
&\textcolor{bestblue}{+.0214}
&\textcolor{abstainred}{$-.0500$}
&.0000
&.0000\\

Sachs
&.0000
&\textcolor{abstainred}{$-.0004$}
&\textcolor{abstainred}{$-.0396$}
&\textcolor{bestblue}{+.0125}\\

Child
&\textcolor{abstainred}{$-.0091$}
&\textcolor{abstainred}{$-.0106$}
&\textcolor{bestblue}{+.0160}
&.0000\\

Insurance
&\textcolor{abstainred}{$-.0088$}
&\textcolor{abstainred}{$-.0057$}
&\textcolor{bestblue}{+.0138}
&\textcolor{bestblue}{+.0304}\\

Alarm
&\textcolor{bestblue}{+.0801}
&\textcolor{bestblue}{+.0089}
&\textcolor{abstainred}{$-.0092$}
&\textcolor{bestblue}{+.0369}\\

\midrule
\multicolumn{5}{l}{\textbf{F1+F2 minus F1+F2-data}}\\
Asia
&\textcolor{abstainred}{$-.0036$}
&\textcolor{abstainred}{$-.0250$}
&.0000
&.0000\\

Sachs
&.0000
&\textcolor{bestblue}{+.0901}
&\textcolor{bestblue}{+.0258}
&\textcolor{bestblue}{+.0625}\\

Child
&.0000
&.0000
&\textcolor{bestblue}{+.0160}
&\textcolor{bestblue}{+.0160}\\

Insurance
&\textcolor{bestblue}{+.0153}
&\textcolor{bestblue}{+.0385}
&\textcolor{bestblue}{+.0133}
&\textcolor{abstainred}{$-.0004$}\\

Alarm
&\textcolor{bestblue}{+.0570}
&\textcolor{bestblue}{+.0167}
&\textcolor{bestblue}{+.0105}
&\textcolor{bestblue}{+.0246}\\
\bottomrule
\end{tabular}
\end{table*}

Allocation effects vary in sign. The semantic-resolution equal-network
mean remains positive under every leave-one-network-out omission in the
four settings. This is an influence diagnostic on the same networks.
It does not add independent evidence or establish a positive interval
after every omission. Exact unrounded omission estimates are not
reconstructed from rounded entries.

\begin{figure*}[t]
\centering
\includegraphics[width=\textwidth]{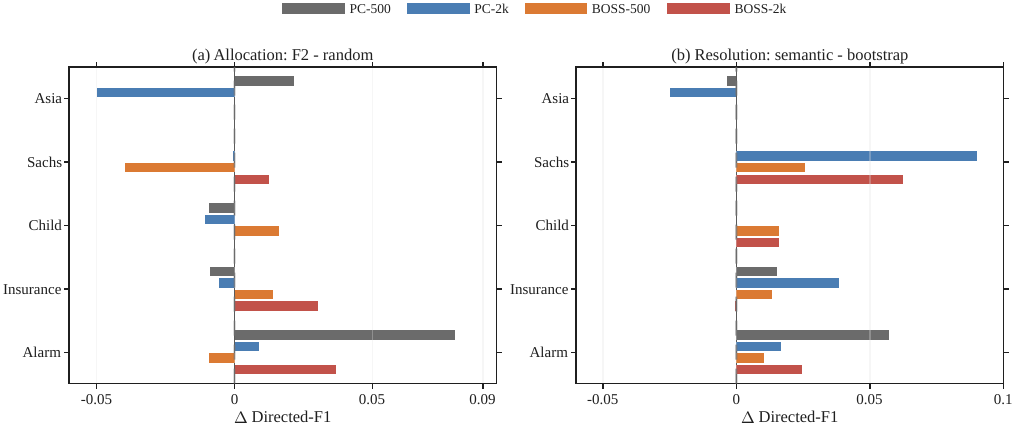}
\caption{\textbf{Network heterogeneity of the F2 contrasts.}
Grouped horizontal bars show
(a) F1+F2 minus matched random and
(b) F1+F2 minus bootstrap-only resolution.
Rows identify networks; colors identify estimator and sample-size settings.
Bars show descriptive within-network Directed-F1 differences,
without network-specific intervals.}
\label{fig:heterogeneity}
\end{figure*}

\subsection{Budget Sweep}

\begin{table*}[t]
\centering\small
\caption{\textbf{Pooled allocation contrasts over response budgets.}
Each cell is $\Delta$Directed-F1 / $\Delta$SHD for F1+F2 minus
F1+rand. Full is a categorical endpoint.}
\label{tab:budget-full}
\begin{tabular}{@{}lrrrr@{}}
\toprule
Cap&PC-500&PC-2000&BOSS-500&BOSS-2000\\
\midrule
0
&.0000 / .00
&.0000 / .00
&.0000 / .00
&.0000 / .00\\

10
&.0000 / .00
&.0000 / .00
&.0000 / .00
&.0000 / .00\\

20
&\textcolor{abstainred}{$-.0057$} / \textcolor{abstainred}{+.04}
&.0000 / .00
&.0000 / .00
&.0000 / .00\\

40
&\textcolor{abstainred}{$-.0067$} / \textcolor{abstainred}{+.08}
&\textcolor{abstainred}{$-.0202$} / \textcolor{abstainred}{+.20}
&\textcolor{abstainred}{$-.0079$} / \textcolor{abstainred}{+.12}
&.0000 / .00\\

80
&\textcolor{bestblue}{+.0054} / \textcolor{bestblue}{$-.08$}
&\textcolor{abstainred}{$-.0116$} / \textcolor{abstainred}{+.15}
&\textcolor{abstainred}{$-.0077$} / \textcolor{abstainred}{+.12}
&\textcolor{bestblue}{+.0056} / \textcolor{bestblue}{$-.08$}\\

Full
&\textcolor{bestblue}{+.0167} / \textcolor{bestblue}{$-.60$}
&\textcolor{abstainred}{$-.0148$} / \textcolor{abstainred}{+.25}
&\textcolor{abstainred}{$-.0038$} / .00
&\textcolor{bestblue}{+.0160} / \textcolor{bestblue}{$-.76$}\\
\bottomrule
\end{tabular}
\end{table*}

\begin{figure}[t]
\centering
\includegraphics[width=\columnwidth]{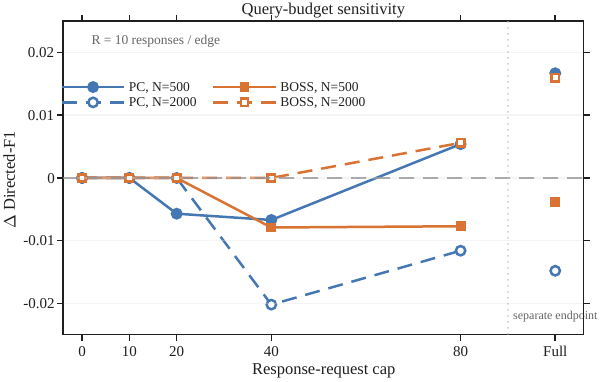}
\caption{\textbf{Allocation across response budgets.}
Pooled paired Directed-F1 differences for F1+F2 minus F1+rand.
Lines connect numerical response-request caps for PC and BOSS at
$N=500,2000$. With $R=10$, caps 10/20/40/80 permit at most
1/2/4/8 queried edges. Full allocation is an unconnected categorical
endpoint. No uncertainty intervals are supplied for this sweep.}
\label{fig:budget}
\end{figure}

The allocation contrast remains non-monotone across response caps and
estimator regimes. Full allocation is kept as a separate endpoint
because candidate counts differ.

\subsection{Additional Families and Protocol Sensitivity}

\begin{table*}[t]
\centering\small
\caption{\textbf{Additional-family and mixed-protocol summaries.}
Values are equal-network Directed-F1 effects. The combined row is a
heterogeneous-protocol sensitivity, not one homogeneous eight-network
experiment.}
\label{tab:expanded}
\begin{tabular}{@{}llrrc@{}}
\toprule
Contrast&Population&Networks&Effect&95\% interval\\
\midrule
F2 minus random
&Primary&5&+.00474&$[-.01527,.02475]$\\
&Additional&3&$-.00104$&$[-.00605,.00398]$\\
&Combined sensitivity&8&+.00257&$[-.00795,.01310]$\\
\midrule
Semantic minus data
&Primary
&5
&\empirical{+.01718}
&$[-.00515,.03951]$\\

&Additional
&3
&\best{+.00418}
&\textcolor{bestblue}{\textbf{$[.00243,.00593]$}}\\

&Combined sensitivity
&8
&\empirical{+.01230}
&$[-.00038,.02499]$\\
\bottomrule
\end{tabular}
\end{table*}

The additional-family semantic-resolution SHD contrast is $-.25146$
$[-.47082,-.03211]$. Its allocation SHD contrast is $+.05166$
$[-.29702,.40033]$. Additional-family recovery has 14, 12, 15, and
15 usable instances per representation in the four settings.
Four PC evaluations are unavailable. Canonical and permuted recovery
results remain separate from the canonical replay summarized above.

% ============================================================
% D. SEMANTIC RELIABILITY
% ============================================================
\section{Semantic Reliability and Representation}
\label{app:semantic}

\begin{table*}[t]
\centering\small
\caption{\textbf{Primary semantic-control counts.}
Raw accuracy and valid-only accuracy use truth-adjacent requests.
UNKNOWN includes 480 nonadjacent controls among 1,760 requests.
Contradiction uses only presentation pairs yielding two parsed
directions. These are different populations.}
\label{tab:semantic-counts}
\begin{tabular}{@{}lrrrr@{}}
\toprule
Condition&Raw correct&Valid-only correct&UNKNOWN&Contradiction\\
\midrule
\textcolor{bestblue}{\textbf{Named}}
&\best{1143/1280}
&\best{1143/1230}
&146/1760
&\best{32/774}\\

No-Benchmark
&1114/1280
&1114/1231
&132/1760
&50/783\\

Anonymous
&1052/1280
&1052/1214
&139/1760
&79/778\\

\textcolor{abstainred}{Shuffled}
&\textcolor{abstainred}{705/1280}
&\textcolor{abstainred}{705/963}
&\textcolor{abstainred}{538/1760}
&\textcolor{abstainred}{95/522}\\
\bottomrule
\end{tabular}
\end{table*}

The conditions are separate interventions, not a nested sequence.
No-Benchmark removes benchmark identity. Anonymous changes variable
identity. Shuffled changes the correspondence between variables and
descriptions. Residual cues can remain. Repeated requests on the same
variables are not independent graph-level observations, so we do not
attach request-binomial confidence intervals to these panels.

At the frozen 80-response cap, Named-minus-Shuffled recovery gains
range from $.0360$ to $.0999$ Directed-F1 across the four primary
settings, with SHD reductions of $.80$--$1.52$. These descriptive
ranges do not establish an uncertainty interval for each paired contrast.

\begin{table*}[t]
\centering\small
\caption{\textbf{Additional-family response counts.}
Each condition--representation cell contains 380 truth-adjacent
requests at temperature $0$. Contradiction has its own paired denominator.}
\label{tab:representation-counts}
\begin{tabular}{@{}llrrrr@{}}
\toprule
Condition&Representation&Raw correct&Valid-only correct&UNKNOWN&Contradiction\\
\midrule
\textcolor{bestblue}{\textbf{Original}}
&Canonical
&\textcolor{bestblue}{305/380}
&\textcolor{bestblue}{305/355}
&\textcolor{bestblue}{25/380}
&\textcolor{bestblue}{20/175}\\

\textcolor{bestblue}{\textbf{Original}}
&Permuted
&\textcolor{bestblue}{316/380}
&\textcolor{bestblue}{316/360}
&\textcolor{bestblue}{20/380}
&\textcolor{bestblue}{24/180}\\

Anonymous
&Canonical
&245/380
&245/335
&45/380
&35/150\\

Anonymous
&Permuted
&242/380
&242/325
&55/380
&31/150\\

\textcolor{abstainred}{Shuffled}
&Canonical
&\textcolor{abstainred}{141/380}
&\textcolor{abstainred}{141/256}
&\textcolor{abstainred}{124/380}
&\textcolor{abstainred}{20/110}\\

\textcolor{abstainred}{Shuffled}
&Permuted
&\textcolor{abstainred}{124/380}
&\textcolor{abstainred}{124/263}
&\textcolor{abstainred}{117/380}
&\textcolor{abstainred}{21/119}\\
\bottomrule
\end{tabular}
\end{table*}

These families are additional to the primary recovery panel, not a
pretraining-disjoint test. Original descriptions remain more accurate
than Anonymous and Shuffled inputs under both representations. The
differences between canonical and permuted values prevent a claim of
exact representation invariance.

% ============================================================
% E. CALIBRATION
% ============================================================
\section{Calibration Feasibility and Empirical Selectors}
\label{app:calibration}

The fixed grid is $\{.5,.6,.7,.8,.9,1.0\}$ and $\eta=.1$.
The real pools contain ten F1 and nine F2 graph tasks. The zero-loss
requirements in Table~\ref{tab:admission} already exceed both pools.
At positive empirical loss $\widehat R<\alpha$, the Hoeffding
requirement becomes
\[
n\ge\left\lceil\frac{\log60}{2(\alpha-\widehat R)^2}\right\rceil.
\]
No finite count makes this bound certify a threshold whose empirical
loss is at least $\alpha$. The zero-loss rows are therefore optimistic
feasibility limits for the stated certificates, not sufficient task
counts under arbitrary observed losses.

\begin{table}[t]
\centering\small
\caption{\textbf{Descriptive empirical selection at $\alpha=.20$.}
Task loss averages $V_g/\max(1,M_g)$ over graph tasks.
Neither selector inherits the tested finite-grid certificate.}
\label{tab:empirical-selectors}
\begin{tabular}{@{}llrr@{}}
\toprule
Policy&Pool&Admission&Task loss\\
\midrule
Add-one
&F1
&\best{1.000}
&\best{.162}\\

Add-one
&F2
&.805
&.175\\

Agreement
&F1
&.944
&\second{.170}\\

Agreement
&F2
&\second{.965}
&.198\\
\bottomrule
\end{tabular}
\end{table}

Add-one uses pooled admitted-edge counts for selection, while the
reported evaluation loss weights graph tasks equally. Agreement uses
the empirical agreement rule. At $\alpha=.10$, the reported F1
Add-one admission fraction is $.858$ with task loss $.070$; the F2
Add-one policy abstains. These observations do not certify target risk.

% ============================================================
% F. FRESH ORACLE
% ============================================================
\section{Fresh Oracle Acquisition and Retention}
\label{app:oracle}

\subsection{Acquisition Protocol}

The Oracle experiment uses Asia, Sachs, and Child, with 80 independent
response-bank draws per network. It requests 15 responses for every
reversible edge in each draw. The model identifier is
\texttt{anthropic/claude-opus-4.5}; transport is OpenRouter with the
Anthropic provider, temperature $.8$, and a 64-token output cap.
Each pair uses a deterministic 8/7 display-order balance, with the
starting side varied by network, draw, and edge.

The system prompt is:
\begin{quote}\small
You are an expert in causal structure. Two variables are related; say
which causes the other. Return ONLY JSON \{"cause": "<name>"\}.
\end{quote}

The user prompt is:
\begin{quote}\small
Between '<first>' and '<second>', which causes the other?
JSON \{"cause":"..."\}.
\end{quote}

The request identity includes protocol, network, draw, edge,
repetition, displayed order, prompt, model, temperature, and output
cap. No fixed network-level vote bank is reused. Invalid or UNKNOWN
responses contribute no vote. At least eight identical valid votes
are required for a proposal. The evaluator scores these proposals
directly; it does not perform sequential structural admission.

All 38,400 planned response slots have a successful provider response.
Of these, 42 do not yield a valid directional parse. The attempt ledger
contains 230,400 provider attempts, including 192,000 attempts beyond
the terminal fingerprints. Attempts and retries are not additional
scientific responses. Completion therefore means successful
acquisition, not 38,400 correct or directionally parsed judgments.

\subsection{References, Effects, and Splits}

The reference is the exact MEC of the true DAG. Exact edge-set lookup
identifies the true member rather than assuming an enumeration index.
The reference sizes are 6 for Asia, 336 for Sachs, and 24 for Child.
Each draw has 1,500 observations. The evaluator selects an
intervention--outcome pair from true ancestor pairs within the draw.
This truth-informed target selection defines a diagnostic population,
not a deployment target-selection procedure.

For each candidate DAG, the evaluator adjusts for the intervention
node's candidate parents. It uses 200 row-bootstrap replicates and
the 1.875th and 98.125th percentiles. These are nominal 96.25\%
member intervals. The retained envelope encloses the member intervals.
The retention level is separately $1-\delta_S=.925$.

The experiment uses 150 repeated random 120/120
calibration--evaluation splits with split seed 7. The rank is
$\lceil121\times.925\rceil=112$. Every split selects $k=4$.
Data-PI keeps the full reference; hard retention keeps
zero-disagreement members; the true-DAG $k$-set uses the calibrated
allowance; size-matched retention draws a random subset with the same
cardinality as the $k$-set. Repeated appearances do not create
additional independent draws or networks.

\subsection{Fixed-$k$ Retention Ablation}
\label{app:oracle-k-sweep}

We vary the disagreement tolerance while keeping the reference,
semantic proposals, target effects, member intervals, and evaluation
splits fixed. The sweep is an ablation of retention only; no value of
$k$ is selected from evaluation performance. Reported values first
average evaluation appearances within each network and then weight the
three networks equally. Width conditions on interval availability.

\begin{table*}[t]
\centering\small
\caption{\textbf{Fresh Oracle fixed-$k$ retention sweep.}
The reference and all downstream quantities are held fixed while only
the allowed semantic-disagreement count changes. The calibrated
procedure selects $k=4$ in all 150 splits. Mean size counts retained
DAGs and includes empty retained sets.}
\label{tab:oracle-k-sweep}
\setlength{\tabcolsep}{5pt}
\begin{tabular}{@{}rrrrrrr@{}}
\toprule
$k$ & Truth & Coverage & Available & Width & Action & Mean size\\
\midrule
0  & .3333 & .5568 & .6859 & .1747 & .6096 & .6859\\
1  & .3333 & .7995 & .8577 & .8250 & .6443 & 5.0026\\
2  & .6859 & .9786 & 1.0000 & .9842 & .7314 & 13.2315\\
3  & .8577 & .9831 & 1.0000 & 1.1113 & .6909 & 26.0454\\

\best{4}
& \best{1.0000}
& \best{.9831}
& \best{1.0000}
& \best{1.1965}
& \best{.6525}
& \best{44.2904}\\

5  & 1.0000 & .9831 & 1.0000 & 1.2382 & .6315 & 64.4527\\
6  & 1.0000 & .9831 & 1.0000 & 1.2555 & .6273 & 84.2676\\
7  & 1.0000 & .9831 & 1.0000 & 1.2633 & .6234 & 101.9768\\
8  & 1.0000 & .9831 & 1.0000 & 1.2633 & .6234 & 114.5296\\
9  & 1.0000 & .9831 & 1.0000 & 1.2633 & .6234 & 120.3807\\
10 & 1.0000 & .9831 & 1.0000 & 1.2633 & .6234 & 121.8534\\
11 & 1.0000 & .9831 & 1.0000 & 1.2633 & .6234 & 122.0000\\
\bottomrule
\end{tabular}
\end{table*}

Hard retention is too restrictive in this truth-containing reference:
at $k=0$, equal-network truth retention is only $.3333$. Increasing
the allowance restores exact truth retention, which reaches $1.0000$
at the calibrated value $k=4$. Further increases retain progressively
more DAGs and eventually recover the Data-PI downstream behavior.
Thus, the calibrated value lies between hard commitment and retaining
the full reference. This sweep is descriptive over three fixed
networks and does not identify an optimal universal value of $k$.

\subsection{Downstream Policy Comparison}

\begin{table*}[t]
\centering\small
\caption{\textbf{Fresh Oracle equal-network results.}
Within-network aggregation weights evaluation appearances. Width is
conditional on interval availability. The true-DAG $k$-set uses the
calibrated value $k=4$.}
\label{tab:oracle-full}
\setlength{\tabcolsep}{5pt}
\begin{tabular}{@{}lrrrrrr@{}}
\toprule
Policy&Truth&Coverage&Available&Width&Action&Act-and-harm\\
\midrule

\second{Data-PI}
&\second{1.0000}
&\second{.9831}
&\second{1.0000}
&\second{1.2633}
&\second{.6234}
&\second{.0000}\\

\textcolor{abstainred}{Hard $k=0$}
&\textcolor{abstainred}{.3333}
&\textcolor{abstainred}{.5568}
&\textcolor{abstainred}{.6859}
&.1747
&.6096
&\textcolor{abstainred}{.0035}\\

\best{True-DAG $k$-set}
&\best{1.0000}
&\best{.9831}
&\best{1.0000}
&\best{1.1965}
&\best{.6525}
&\best{.0000}\\

Size-matched
&.7660
&.9831
&1.0000
&1.2628
&.6234
&.0000\\

\bottomrule
\end{tabular}
\end{table*}

\begin{figure*}[t]
\centering
\includegraphics[width=\textwidth]{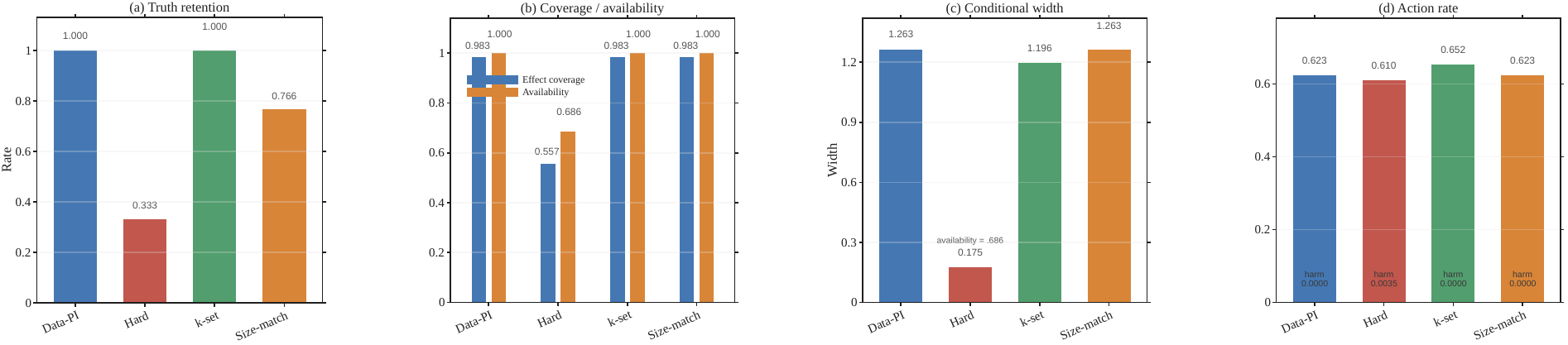}
\caption{\textbf{Fresh Oracle retention under an adequate reference.}
Policies are Data-PI, hard $k=0$, calibrated true-DAG disagreement
$k$-set, and size-matched random retention. Panels show
(a) exact truth retention, (b) effect coverage and interval
availability, (c) width conditional on availability, and
(d) action rate with joint act-and-harm annotated.
Values weight evaluation appearances within each network and then
weight the three networks equally. They are descriptive; no
cross-network intervals are asserted. Member intervals use the
1.875th and 98.125th percentiles (96.25\% nominal level).
Retention calibration uses $1-\delta_S=.925$, a different quantity.}
\label{fig:oracle}
\end{figure*}

\begin{table}[t]
\centering\small
\caption{\textbf{Oracle availability counts.}
Counts are pooled, whereas Table~\ref{tab:oracle-full} reports
equal-network rates. Each policy has 240 draws and 18,000 evaluation
appearances.}
\label{tab:oracle-availability}
\setlength{\tabcolsep}{4pt}
\begin{tabular}{@{}lrr@{}}
\toprule
Policy&Available draws&Available appearances\\
\midrule

\textcolor{bestblue}{Data-PI}
&\textcolor{bestblue}{240}
&\textcolor{bestblue}{18,000}\\

\textcolor{abstainred}{Hard $k=0$}
&\textcolor{abstainred}{165}
&\textcolor{abstainred}{12,388}\\

\best{True-DAG $k$-set}
&\best{240}
&\best{18,000}\\

\textcolor{bestblue}{Size-matched}
&\textcolor{bestblue}{240}
&\textcolor{bestblue}{18,000}\\

\bottomrule
\end{tabular}
\end{table}

The hard-policy conditional width $.1747$ differs from its zero-coded
score $.1294$. Its equal-network availability $.6859$ differs from
the pooled appearance rate because network weights differ. No interval
over all 18,000 appearances is treated as an independent-draw or
independent-network confidence interval.

\subsection{Per-Network Behavior}

\begin{table*}[t]
\centering\small
\caption{\textbf{Fresh Oracle behavior by network.}
These entries average the 80 draw-level summaries equally within each
network. They therefore use a different within-network weighting from
Table~\ref{tab:oracle-full}.}
\label{tab:oracle-network}
\begin{tabular}{@{}llrrrrrr@{}}
\toprule
Network&Policy&Truth&Coverage&Available&Width&Action&Mean size\\
\midrule

Asia&Data-PI&1.0000&1.0000&1.0000&.7479&.7875&6.0\\
&Hard&1.0000&.9875&1.0000&.2078&1.0000&1.0\\
&\best{$k$-set}
&\best{1.0000}
&\best{1.0000}
&\best{1.0000}
&\best{.7479}
&\best{.7875}
&\best{6.0}\\
&Size-matched&1.0000&1.0000&1.0000&.7479&.7875&6.0\\

\midrule

Child&Data-PI&1.0000&.9500&1.0000&1.1671&.5750&24.0\\
&\textcolor{abstainred}{Hard}
&\textcolor{abstainred}{.0000}
&\textcolor{abstainred}{.6750}
&1.0000
&.1756
&.7750
&1.0\\
&\best{$k$-set}
&\best{1.0000}
&\best{.9500}
&\best{1.0000}
&\best{1.1671}
&\best{.5750}
&\best{24.0}\\
&Size-matched&1.0000&.9500&1.0000&1.1671&.5750&24.0\\

\midrule

Sachs&Data-PI&1.0000&1.0000&1.0000&1.8846&.5125&336.0\\
&\textcolor{abstainred}{Hard}
&\textcolor{abstainred}{.0000}
&\textcolor{abstainred}{.0125}
&\textcolor{abstainred}{.0625}
&.1436
&\textcolor{abstainred}{.0625}
&.0625\\
&\best{$k$-set}
&\best{1.0000}
&\best{1.0000}
&\best{1.0000}
&\best{1.6837}
&\best{.6000}
&\best{103.0625}\\
&Size-matched
&\textcolor{abstainred}{.2989}
&1.0000
&1.0000
&1.8831
&.5127
&103.0625\\

\bottomrule
\end{tabular}
\end{table*}

The calibrated $k$-set leaves Asia and Child unchanged relative to
Data-PI. Its pruning, narrowing, and action changes occur in Sachs.
The size-matched policy can lose the exact graph while retaining
empirical effect coverage. Graph retention and effect coverage are
therefore distinct outcomes even within an adequate Oracle reference.

% ============================================================
% G. LEARNED X6 AND REFERENCE ABLATIONS
% ============================================================
\section{Learned References and the X6 Diagnostic}
\label{app:x6}

\subsection{Population and Retention Definitions}

Exact-MEC enumeration yields 72 downstream-usable tasks from 80 planned
tasks. Three tasks reach the enumeration limit and five have no valid
MEC extension. Truth is present in three usable references. Planned-task
coverage counts failed construction as absence:
$3/80=.0375$. Among downstream-usable tasks, exact membership is
$3/72=.0417$.

These counts refer to the historical downstream X6 enumeration and
usability stage. Operationally, that downstream cohort excludes three
tasks whose exact-MEC enumeration reaches the enumeration limit and
five tasks for which no valid MEC extension is obtained. The separate
construction ablation in Table~\ref{tab:x6-reference-construction}
records construction outputs over all 80 planned tasks and does not
apply this downstream usability criterion. Thus, a construction record
being available does not imply that the corresponding task is usable
by the downstream X6 evaluator.

The closest member minimizes directed-edge symmetric-difference distance
to truth within the stored reference, with first-member tie-breaking.
The historical reference-disagreement policy calibrates against this
member. It is a truth-informed diagnostic proxy, not a deployable
substitute for true-DAG calibration. High closest-member retention does
not establish exact truth retention when truth is absent upstream.

\subsection{Two Calibration Diagnostics}

The downstream policy evaluation uses 120 repeated 36/36 splits of the
72 usable tasks, yielding 4,320 evaluation appearances per policy.
Its reference-disagreement rank at $\delta_S=.05$ is
$\lceil37\times.95\rceil=36$.

The separate theorem-aligned diagnostic specifies $m=40$,
$\delta_S=.05$, and 120 splits; its rank is 39. These protocols define
different evaluation populations and are not pooled.

For the true-DAG diagnostic, selected $k$ is 2 in 58 splits and 3 in
62 splits. Marginal post-retention truth coverage is $.0395833$ over
its evaluation appearances. Conditional survival is $1.0000$, but only
three distinct tasks contain truth upstream. Repeated evaluation
appearances do not enlarge that sample.

The corresponding separate reference-disagreement diagnostic selects
$k=2$ in 13 splits and $k=3$ in 107 splits. It also reports
$.0395833$ marginal truth retention and $1.0000$ conditional survival.
These diagnostic summaries do not provide downstream effect results for
the true-DAG scoring policy.

\subsection{Downstream Policy Comparison}

The historical downstream evaluator uses reference-disagreement scoring.
All entries below use the 72 downstream-usable tasks and 4,320
evaluation appearances per policy. Width is conditional on interval
availability; coverage, action, and joint act-and-harm use all
appearances.

\begin{table*}[t]
\centering\small
\caption{\textbf{Learned X6 downstream diagnostic.}
Closest denotes retention of the truth-informed closest stored member,
not exact truth retention. The displayed $k$-set calibrates
reference-member disagreement. Oracle results are not pooled with this
population.}
\label{tab:x6}
\setlength{\tabcolsep}{5pt}
\begin{tabular}{@{}lrrrrrr@{}}
\toprule
Policy&Closest&Available&Width&Coverage&Action&Act-and-harm\\
\midrule

\second{Data-PI}
&\second{1.0000}
&\second{1.0000}
&\second{.8227}
&\second{.6028}
&\second{.7963}
&\second{.0132}\\

\textcolor{abstainred}{Hard $k=0$}
&\textcolor{abstainred}{.4563}
&\textcolor{abstainred}{.6637}
&.3269
&\textcolor{abstainred}{.4361}
&.5896
&.0123\\

\best{Reference-disagreement $k$-set}
&\best{.9965}
&\best{1.0000}
&\best{.8226}
&\best{.6028}
&\best{.7963}
&\best{.0132}\\

Size-matched
&.8192
&1.0000
&.8219
&.6012
&.7963
&.0132\\

Data-only
&.4477
&1.0000
&.4728
&.5181
&.8618
&.0132\\

\bottomrule
\end{tabular}
\end{table*}

The calibrated reference-disagreement policy retains the closest member
with frequency $.9965$, but its effect coverage remains $.6028$.
This result concerns preservation of an available proxy. It does not
repair the low exact-truth membership of the stored reference.

\subsection{Fixed-$k$ Retention Sensitivity}
\label{app:x6-k-sweep}

We next vary only the allowed reference-disagreement count. The stored
references, semantic score lists, effects, member intervals, and
evaluation splits remain fixed. Fixed values of $k$ are not selected
using downstream performance.

\begin{table*}[t]
\centering\small
\caption{\textbf{Learned-reference fixed-$k$ sweep.}
All downstream columns are pooled over the 4,320 evaluation appearances.
Exact is exact true-DAG retention in this same appearance-weighted
population; Closest is retention of the truth-informed closest stored
member. Width conditions on availability. The calibrated row uses the
split-specific reference-disagreement allowance rather than one fixed
$k$.}
\label{tab:x6-k-sweep}
\setlength{\tabcolsep}{4.5pt}
\begin{tabular}{@{}lrrrrrrr@{}}
\toprule
Policy
&\textcolor{abstainred}{Exact}
&Closest
&Available
&Coverage
&Width
&Action
&Mean size\\
\midrule

$k=0$
&\textcolor{abstainred}{.0428}
&.4563&.6637&.4361&.3269&.5896&.6898\\

$k=1$
&\textcolor{abstainred}{.0428}
&.7074&.9162&.5343&.7917&.7387&4.1417\\

$k=2$
&\textcolor{abstainred}{.0428}
&.9322&1.0000&.6028&.8169&.7963&10.7796\\

$k=3$
&\textcolor{abstainred}{.0428}
&1.0000&1.0000&.6028&.8227&.7963&18.7037\\

$k=4$
&\textcolor{abstainred}{.0428}
&1.0000&1.0000&.6028&.8227&.7963&25.8238\\

$k=5$
&\textcolor{abstainred}{.0428}
&1.0000&1.0000&.6028&.8227&.7963&30.6512\\

$k=6$
&\textcolor{abstainred}{.0428}
&1.0000&1.0000&.6028&.8227&.7963&33.0639\\

$k=7$
&\textcolor{abstainred}{.0428}
&1.0000&1.0000&.6028&.8227&.7963&33.9495\\

$k=8$
&\textcolor{abstainred}{.0428}
&1.0000&1.0000&.6028&.8227&.7963&34.1715\\

$k=9$
&\textcolor{abstainred}{.0428}
&1.0000&1.0000&.6028&.8227&.7963&34.1993\\

\midrule

\best{Calibrated $k$-set}
&\textcolor{abstainred}{\textbf{.0428}}
&\best{.9965}
&\best{1.0000}
&\best{.6028}
&\best{.8226}
&\best{.7963}
&\best{18.4759}\\

\second{Data-PI}
&\textcolor{abstainred}{.0428}
&\second{1.0000}
&\second{1.0000}
&\second{.6028}
&\second{.8227}
&\second{.7963}
&\second{34.1993}\\

\bottomrule
\end{tabular}
\end{table*}

Hard retention at $k=0$ already matches Data-PI exact truth retention
at the reported precision. Increasing $k$ preserves this value while
restoring additional stored reference members: closest-member retention
reaches $1.0000$ at $k=3$, and availability reaches $1.0000$ at $k=2$.
Pruning cannot recover a true DAG that is absent from the stored reference.

The $.0428$ quantity is specific to the 4,320-appearance downstream
population above. It is not the $.0395833$ marginal truth-retention
estimand from the separate theorem-aligned diagnostic in the previous
subsection. Neither quantity replaces the task-level reference coverage
$3/72=.0417$.

\subsection{Availability and Retained-Set Size}

The historical X6 evaluator uses 120 effect-bootstrap replicates and
nominal 95\% member intervals. Its protocol differs from the fresh
Oracle experiment in reference construction, proposal handling, target
population, and member-interval level. Absolute width and coverage
values are therefore not interpreted as an Oracle--X6 treatment
contrast.

\begin{table*}[t]
\centering\small
\caption{\textbf{X6 availability and pruning.}
Each policy has 72 usable tasks and 4,320 evaluation appearances.
Nonempty-task counts and available-appearance counts have different
units. Mean shrinkage is averaged per task or appearance rather than
computed from aggregate sizes. Zero-coded width assigns zero when an
interval is unavailable and is distinct from conditional width.}
\label{tab:x6-availability}
\setlength{\tabcolsep}{5pt}
\begin{tabular}{@{}lrrrrr@{}}
\toprule
Policy
&Nonempty tasks
&Available appearances
&Mean size
&Mean shrinkage
&Zero-coded width\\
\midrule

\second{Data-PI}
&\second{72}
&\second{4320}
&\second{34.1993}
&\second{.0000}
&\second{.8227}\\

\textcolor{abstainred}{Hard $k=0$}
&\textcolor{abstainred}{48}
&\textcolor{abstainred}{2867}
&.6898
&.9283
&\textcolor{abstainred}{.2169}\\

\best{Reference-disagreement $k$-set}
&\best{72}
&\best{4320}
&\best{18.4759}
&\best{.1776}
&\best{.8226}\\

Size-matched
&72
&4320
&18.4759
&.1776
&.8219\\

Data-only
&72
&4320
&8.7106
&.6207
&.4728\\

\bottomrule
\end{tabular}
\end{table*}

The calibrated reference-disagreement policy reduces mean retained-set
size from $34.1993$ to $18.4759$, but its downstream envelope and
decisions remain nearly unchanged. This reduction concerns the stored
reference only; it does not improve upstream exact-truth membership.

\subsection{Reference-Construction Ablation}

We also broaden the stored reference independently of the retention
rule. This is a separate construction-stage ablation over all 80 planned
tasks, not the 72-task downstream X6 population defined above.
Coverage uses all 80 planned tasks, and size summaries condition on
construction records available for the corresponding method.

\begin{table*}[t]
\centering\small
\caption{\textbf{Reference construction on 80 planned tasks.}
Records denotes construction outputs produced by this ablation.
Record failures are construction-stage failures and do not include the
separate downstream enumeration-limit or valid-extension exclusions.
Planned coverage uses all 80 tasks. Mean and median size condition on
produced construction records.}
\label{tab:x6-reference-construction}
\setlength{\tabcolsep}{5pt}
\begin{tabular}{@{}lrrrrr@{}}
\toprule
Construction
&Records
&Record failures
&Planned coverage
&Mean size
&Median size\\
\midrule

\second{Exact MEC}
&\second{80}
&\second{0}
&\second{.0375}
&\second{44.5625}
&\second{12}\\

Skeleton-relaxed
&74
&\textcolor{abstainred}{6}
&\textcolor{bestblue}{.0500}
&280.3919
&61\\

Bootstrap union
&78
&\textcolor{abstainred}{2}
&\best{.0750}
&\textcolor{abstainred}{4291.4487}
&\textcolor{abstainred}{116}\\

\bottomrule
\end{tabular}
\end{table*}

Broader construction improves exact-truth coverage only modestly.
The bootstrap union raises planned coverage to $.0750$, but mean stored
size increases to $4291.4487$. Its mean and median also differ strongly,
which indicates a skewed size distribution. More stored graphs therefore
do not by themselves establish an adequate reference. These construction
variants do not supply a new matched downstream effect comparison.
% ============================================================
% H. ROBUSTNESS AND IMPLEMENTATION SENSITIVITY
% ============================================================
\section{Robustness and Implementation Sensitivity}
\label{app:robustness}

\subsection{Instability and Adjacency Support}
\begin{table}[t]
\centering\small
\caption{\textbf{Instability-threshold sensitivity.}
Wrong recall concerns incorrectly oriented fitted arrows. Correct
queried is the fraction of already-correct arrows selected.}
\label{tab:tau}
\begin{tabular}{@{}rrrr@{}}
\toprule
$\tau$&Candidates&Wrong recall&Correct queried\\
\midrule
.50&0&.00&.000\\
.60&29&.32&.165\\
.70&57&.70&.278\\

\best{.75}
&\best{68}
&\best{.80}
&\best{.354}\\

.80&80&.90&.443\\
.90&97&.98&.608\\
.95&107&1.00&.722\\
\bottomrule
\end{tabular}
\end{table}

\begin{table*}[t]
\centering\small
\caption{\textbf{Adjacency-support sensitivity.}
Compelled, reversible, and false-adjacency fractions describe the
selected candidate pool. Higher compelled fraction does not imply
higher recovery.}
\label{tab:support}
\begin{tabular}{@{}rrrrrr@{}}
\toprule
Support&Candidates&Compelled&Reversible&False adjacency&Mean Directed-F1\\
\midrule

\second{.00}
&\second{360}
&.511
&.319
&\textcolor{abstainred}{.169}
&\best{.684}\\

.25
&287
&.620
&.380
&\textcolor{bestblue}{.000}
&.681\\

.50
&167
&.689
&.311
&\textcolor{bestblue}{.000}
&.663\\

.75
&52
&.923
&.077
&\textcolor{bestblue}{.000}
&\textcolor{abstainred}{.639}\\

\bottomrule
\end{tabular}
\end{table*}
\subsection{Bootstrap-Count Analysis}
\begin{table*}[t]
\centering\small
\caption{\textbf{Five-network bootstrap diagnostic.}
Candidate means and equal-network Jaccard weight networks equally.
Pooled Jaccard concerns the pooled edge-set comparison with $B=250$.
Semantic and matched-random recovery at these bootstrap counts are
unavailable because the corresponding response records were not
verified for this analysis.}
\label{tab:bootstrap-corrected}
\begin{tabular}{@{}rrrrrr@{}}
\toprule
$B$&Mean candidates&Pooled Jaccard&Equal-network Jaccard&Mean support&Wrong committed\\
\midrule

\best{25}
&\best{3.4}
&\best{.737}
&\best{.7933}
&\best{.8827}
&\best{9}\\

50
&3.2
&.778
&.6500
&.9093
&8\\

100
&3.2
&.778
&.6500
&.8982
&7\\

\second{250}
&\second{3.2}
&\second{1.000}
&\second{1.0000}
&\second{.9585}
&\second{8}\\

\bottomrule
\end{tabular}
\end{table*}

The equal-network overlap is not monotone in this diagnostic.
The value at $B=250$ is one by comparison with itself.
A separate six-network overlap sequence is not combined with this table.
No inference about improved allocation recovery follows from candidate
overlap alone.

\begin{figure*}[t]
\centering
\includegraphics[width=\textwidth]{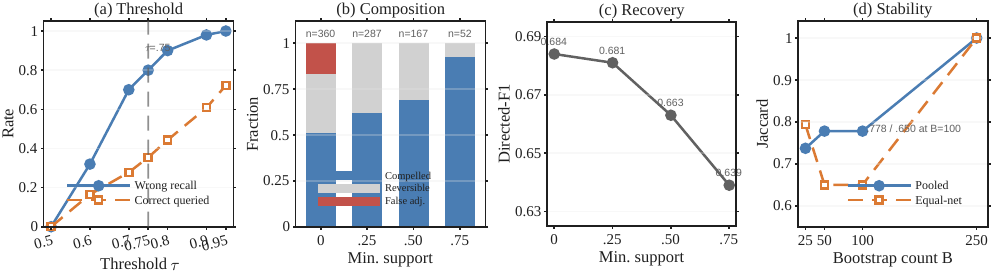}
\caption{\textbf{Locator sensitivity measures different properties.}
(a) Wrong-arrow recall and already-correct-arrow query rate across
instability thresholds.
(b) Candidate composition across adjacency-support thresholds.
(c) Mean Directed-F1 across those support thresholds.
(d) Pooled and equal-network candidate-set Jaccard relative to
$B=250$.
The sweeps use distinct populations.
Candidate overlap does not establish a recovery benefit at larger
bootstrap counts.}
\label{fig:locator-diagnostics}
\end{figure*}

\subsection{Order and Completion}
Reversing canonical proposal order changes recovery in 27 of 95
matched feasible instances. Mean Directed-F1 changes by $-.0192$
and SHD by $+.495$. Proposal order is therefore part of the fixed
policy, not an invariant operation.

The current completion audit agrees with exhaustive enumeration in
20 of 20 small cases, with zero observed false accepts and false
rejects. A forced timeout returns unknown. These checks validate the
tested cases. They do not prove completeness for all relaxed
constraint classes or truth of semantic proposals.

A matched comparison of $R=10$ and $R=15$ is unavailable.
The primary and Oracle response banks differ in task population and
procedure. We do not interpret their difference as a response-count
ablation.

% ============================================================
% I. EXTERNAL METHOD CONTEXT
% ============================================================
\section{Native Guess2Graph Context}
\label{app:guess2graph}

Guess2Graph uses a whole-DAG expert guess to guide statistical
discovery. The native comparison uses the official discovery
implementation, Fisher-$z$ tests at $.05$, and the same frozen PC
data for its compatible comparison. Predicted graphs are symmetrized
before scoring. This is a skeleton-recovery comparison, not a
matched-budget test of pairwise semantic orientation.
Table~\ref{tab:guess2graph} reports both absolute paired metrics and
their differences from the shared PC baseline.

\begin{table*}[t]
\centering\small
\caption{\textbf{Paired native Guess2Graph skeleton comparison.}
Absolute metrics are computed on the paired usable PC-family instances.
Differences are Guess2Graph minus the shared PC baseline. This is a
native skeleton-recovery comparison, not a matched-budget directional
comparison.}
\label{tab:guess2graph}
\setlength{\tabcolsep}{5pt}
\begin{tabular}{@{}rrrrrrrr@{}}
\toprule
$N$
& Paired $n$
& PC Skel.-F1
& G2G Skel.-F1
& $\Delta$Skel.-F1
& PC Adj.-SHD
& G2G Adj.-SHD
& $\Delta$Adj.-SHD\\
\midrule

500
&19
&.7963
&\best{.8217}
&\textcolor{bestblue}{+.0254}
&12.632
&\best{12.263}
&\textcolor{bestblue}{$-.368$}\\

2000
&16
&.8342
&\best{.8485}
&\textcolor{bestblue}{+.0142}
&12.375
&\best{11.625}
&\textcolor{bestblue}{$-.750$}\\

\bottomrule
\end{tabular}
\end{table*}

There are 50 requested instances, 38 usable native outputs, and 35
paired evaluations. Whole-graph requests are not equated to pairwise
orientation requests. The acquisition record reports 119 lifetime
attempts, 62 linked retries, and 69 attempts beyond the 50 initial
requests. These counters describe different aspects of the history.
A parser-correction restart reset a per-request counter, so no strict
three-attempt lifetime cap is claimed. Ten preserved responses across
seven request fingerprints were recovered offline. None of these
acquisition events creates another scientific evaluation instance.

Earlier MosaCD-style and fixed-skeleton Guess2Graph-style adaptations
are not treated as native method evaluations. The present paper makes
no state-of-the-art claim based on those superseded tables.

\end{document}